\documentclass[10pt,twocolumn,letterpaper]{article}

\usepackage[pagenumbers]{iccv} 

\definecolor{iccvblue}{rgb}{0.21,0.49,0.74}
\usepackage[pagebackref,breaklinks,colorlinks,allcolors=iccvblue]{hyperref}
\usepackage{algorithm}
\usepackage{algorithmic}

\usepackage{multirow}

\def\paperID{5115} 
\def\confName{ICCV}
\def\confYear{2025}

\title{Test-Time Discovery via Hashing Memory}

\author{Fan Lyu$^{1*}$, Tianle Liu$^{2*}$, Zhang Zhang$^1$, Fuyuan Hu$^2$, Liang Wang$^1$\\
$^1$ New Laboratory of Pattern Recognition, Institute of Automation, Chinese Academy of Sciences\\
$^2$ School of Electric \& Information Engineering, Suzhou University of Science and Technology\\
{\tt\small fan.lyu@cripac.ia.ac.cn, tianleliu@post.usts.edu.cn, zzhang@nlpr.ia.ac.cn}
\thanks{Equal contributions.}
}

\def\app{\textcolor{purple}}

\def\mb{\mathbf}
\def\mbb{\mathbb}
\def\mc{\mathcal}

\def\ie{\textit{i.e.}}

\begin{document}
\maketitle
\begin{abstract}
We introduce Test-Time Discovery (TTD) as a novel task that addresses class shifts during testing, requiring models to simultaneously identify emerging categories while preserving previously learned ones. A key challenge in TTD is distinguishing newly discovered classes from those already identified. 
To address this, we propose a training-free, hash-based memory mechanism that enhances class discovery through fine-grained comparisons with past test samples. Leveraging the characteristics of unknown classes, our approach introduces hash representation based on feature scale and directions, utilizing Locality-Sensitive Hashing (LSH) for efficient grouping of similar samples.
This enables test samples to be easily and quickly compared with relevant past instances.
Furthermore, we design a collaborative classification strategy, combining a prototype classifier for known classes with an LSH-based classifier for novel ones. To enhance reliability, we incorporate a self-correction mechanism that refines memory labels through hash-based neighbor retrieval, ensuring more stable and accurate class assignments.
Experimental results demonstrate that our method achieves good discovery of novel categories while maintaining performance on known classes, establishing a new paradigm in model testing.
Our code is available at \url{https://github.com/fanlyu/ttd}.
\end{abstract}    
\section{Introduction}
\label{sec:intro}

\begin{figure}[t]
\centering
\begin{subfigure}{\linewidth}
    \centering
    \includegraphics[width=1\linewidth]{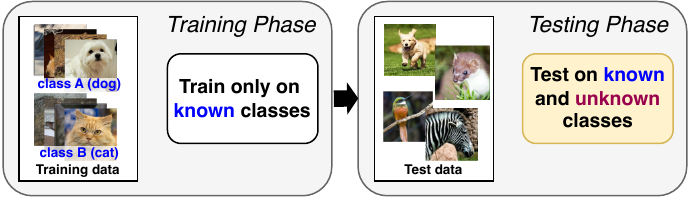}
    \caption{Test-Time Discovery (TTD).}
    \label{fig:ttd_a}
\end{subfigure}
\centering
\begin{subfigure}{\linewidth}
    \centering
    \includegraphics[width=1\linewidth]{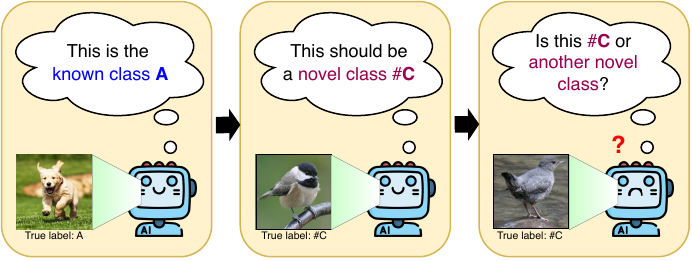}
    \caption{Testing phase: Prediction and Discovery}
    \label{fig:ttd_b}
\end{subfigure}
\vspace{-20px}
\caption{Test-Time Discovery (TTD). A model is initially trained on data containing only known classes. During deployment, the test data may include both known and unknown classes, requiring the model to predict known classes and discover novel ones.
Given a test sample, the model needs to determine whether it belongs to a newly discovered unseen class or an already identified seen class. The small scale of the discovered data and insufficient learning make this distinction challenging.}
\label{fig:ttd}
\end{figure}

In recent years, Test-Time Training (TTT)~\cite{wangtent,boudiaf2022parameter} has emerged as a promising approach, focusing on adapting pre-trained models to domain changes during inference. However, TTT methods typically emphasize predefined categories and overlook the discovery of novel categories at test time.
On the other hand, Novel Category Discovery (NCD)~\cite{zhong2021neighborhood,vaze2022generalized} has made significant advances in identifying new categories without supervision, usually through post-clustering in offline settings. While effective in such contexts, NCD methods are not suitable for real-time test scenarios, where models must dynamically discover and adapt to new categories. 
This paper introduces \textbf{Test-Time Discovery} (TTD), a novel task where a model, initially trained on known classes, is tested online with both known and unknown classes. 
As shown in Fig.~\ref{fig:ttd_a}, a TTD model must predict either a known class or assign a new one. This task is particularly urgent in fields like healthcare, where a diagnostic model must recognize new pathologies during inference, or in autonomous vehicles, which might encounter unexpected obstacles, such as livestock on rural roads.

TTD focuses on class shifts rather than domain shifts during test time, requiring the model to simultaneously discover and classify new classes while maintaining performance on old ones. However, this is challenging due to the complexity of class discovery, limited labeled data, and ambiguous class boundaries. 
As shown in Fig.~\ref{fig:ttd_b}, key challenges include: (1) distinguishing between new class discovery and identification, (2) learning new classes with insufficient samples, and (3) the risk of catastrophic forgetting, where learning new classes can degrade performance on previously learned ones.
Traditional TTT and NCD methods often rely on simple confidence-based thresholding when encountering new classes. 
However, this approach struggles to address key challenges, particularly in distinguishing between already discovered and newly discovered classes, leading to redundant or false class identifications. 
To address these issues, a more refined comparison with similar samples from past test instances could help improve class identification and reduce ambiguity.

In this paper, we propose a training-free, hash-based memory mechanism for the TTD task, which enhances prediction and new class discovery by incorporating memory comparisons with past test samples. Specifically, we leverage insights from previous work on the feature characteristics of known and unknown class samples to design a hash representation method based on feature scale and direction. To improve efficiency, we employ Locality-Sensitive Hashing (LSH) to store test samples' features with similar hash characteristics in the same memory bucket. During testing, samples can quickly locate approximate neighbors within the same or adjacent buckets using their hash values, facilitating more refined comparisons. Additionally, recognizing that the prototype classifier performs more reliably for known classes, we adopt a collaborative classification approach that combines the prototype classifier and the LSH classifier. Finally, since stored samples may provide inaccurate classification guidance due to erroneous pseudo-labels, we introduce a self-correction method for memory. This method uses hash-based neighbor retrieval to assess the reliability of labels in memory and correct them, thus enhancing the overall reliability of the memory. Extensive experiments show that our approach not only outperforms existing NCD methods in test-time scenarios but also achieves robust clustering and classification of novel categories.

Our contributions are three-fold:

\begin{enumerate}[label=(\arabic*),left=0pt,itemsep=0pt]
    \item To the best of our knowledge, we are the first to study the task of novel category discovery at test time.  
    \item We propose a novel training-free mechanism for TTD, which builds memory buffers based on LSH.
    \item We propose a memory self-correction mechanism for relabeling samples in the memory buffer.
\end{enumerate}

\section{Related Work}
\label{sec:rel}


\label{sec:rel_ttl}

\noindent
\textbf{Test-Time Training} (TTT)~\cite{sun2020test,liu2021ttt++,gandelsman2022test,osowiechi2023tttflow,su2024revisiting} is a machine learning approach where a model updates its parameters during inference using auxiliary self-supervised tasks. These tasks leverage input data structures to improve adaptability to distribution shifts, enabling TTT to refine feature extractors and enhance performance on out-of-distribution data.
Test-Time Adaptation (TTA)~\cite{wangtent,liang2020we,boudiaf2022parameter,chen2022contrastive,yuan2023robust,lyu2024variational,lyu2025conformal,shi2024controllable,tian2024parameter,ni2025maintain}, in contrast, adjusts a model dynamically during inference without explicit training. It typically updates lightweight parameters, such as batch normalization statistics, or employs simple strategies like entropy minimization to adapt to new data distributions. Designed for real-time applications, TTA enhances robustness with minimal computational overhead, making it effective for handling mild distribution shifts.
However, TTT focuses solely on distribution shifts and overlooks scenarios where new classes emerge during testing. In open-world environments, models must not only adapt to distribution changes but also autonomously discover and classify novel categories, improving their adaptability to unforeseen challenges.

\noindent
\textbf{Novel Category Discovery} (NCD)~\cite{han2019learning,han2020automatically,zhao2021novel,vaze2022generalized,zhang2023promptcal,cendra2025promptccd} enables models to identify and adapt to previously unseen classes during inference, particularly in open-world settings. It allows models to detect new categories without labeled data, ensuring novel class representation while preserving prior knowledge. Unlike traditional classification, NCD does not require immediate labeling; instead, it clusters additional test data for post-hoc evaluation. However, this reliance on offline clustering makes NCD unsuitable for test-time applications, where immediate classification and seamless integration of new discoveries are essential.

\section{Test-Time Discovery}
\label{sec:ttd}

\subsection{Problem Definition}

In TTD, the model not only needs to classify known categories but also dynamically discover new categories during the testing phase. 
To avoid confusion, we follow the NCD task and categorize all test samples into three groups: known, seen, and unseen. ``\textbf{Known}'' refers to previously trained classes, ``\textbf{seen}'' denotes discovered classes, and ``\textbf{unseen}'' represents undiscovered classes.
As shown in Fig.~\ref{fig:ttd_a}, in the training phase, the model is trained on the known class data  $\mathcal{D}^\text{train}_\text{known} = \{(x_i, y_i)\}_{i=1}^{N^\text{train}} $ to obtain the trained model $f(\cdot)$, where $x_i$ is a data point and $y_i \in \mathcal{Y}_\text{known}$ represents the corresponding known class label.
In the testing phase, the model is tested on a dataset $\mathcal{D}^\text{test} = \mathcal{D}^\text{test}_\text{known} \cup \mathcal{D}^\text{test}_\text{unknown}$, where no true label is given.
A TTD model should achieve two objectives.
First, the model needs to discover unseen categories for each input $x \in \mathcal{D}^\text{test}_\text{unknown}$ with undiscovered category and generate a new category (e.g. ``new class \#C1'') or assign by human to the discovered class set $ \mathcal{Y}_\text{seen}$.
Second, for each sample $x \in \mathcal{D}^\text{test}$, it needs to correctly classify known and seen categories $y \in \mathcal{Y}^\text{test} = \mathcal{Y}_\text{known} \cup \mathcal{Y}_\text{seen}$ if the corresponding category is not undiscovered.

\subsection{The Challenge of TTD}

\label{sec:challenge}

{TTD focuses on class shifts rather than domain shifts}, and the challenge of discovering new classes is more complex than domain adaptation in test-time scenarios. This complexity arises from the limited availability of labeled data for new classes and the ambiguity of class boundaries.
We here list the challenges of the task of TTD.
\textit{(1) New class discovery vs. new class identification}:
When a test sample is recognized as belonging to a novel category, it is crucial to determine whether it aligns with an already discovered class.
\textit{(2) Discovered new classes are hard to learn}:
Once a new class is identified, the limited number of samples makes it insufficient to train the model directly on these classes, often leading to overfitting or underfitting.
\textit{(3) Forgetting of known classes}:
The process of discovering and learning new classes can negatively impact the performance of previously known classes, leading to a decline in accuracy and causing catastrophic forgetting~\cite{kirkpatrick2017overcoming}.

\begin{figure}[t]
\centering
\begin{minipage}{\linewidth}
\begin{subfigure}{0.455\linewidth}
    \centering
    \includegraphics[width=1\linewidth]{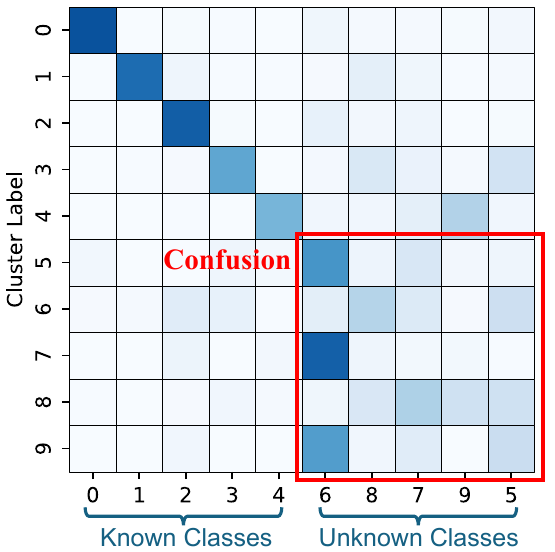}
    \vspace{-15px}
    \caption{Thresholding method}
    \label{fig:mnist_euc}
\end{subfigure}
\hfill
\begin{subfigure}{0.525\linewidth}
    \centering
    \includegraphics[width=1\linewidth]{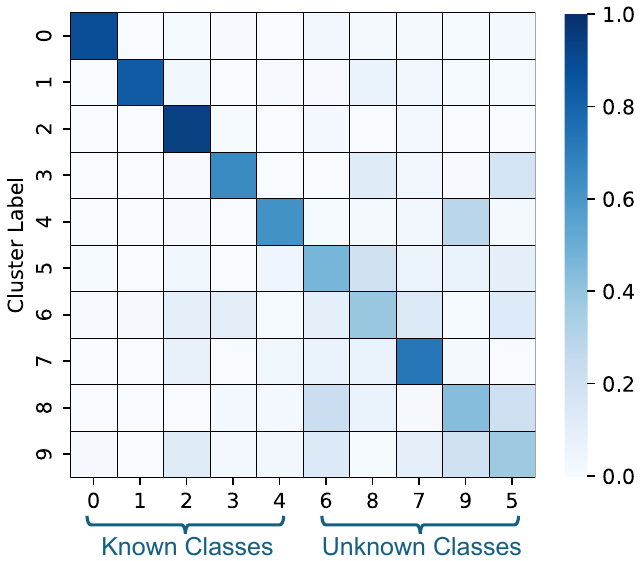}
    \vspace{-15px}
    \caption{Ours}
    \label{fig:mnist_ours}
\end{subfigure}
\end{minipage}
\vspace{-10px}
\caption{
Class-cluster prediction matching matrix on MNIST ($5+5$).
For each prediction cluster, we compute the class composition and visualize the relationships. To enhance clarity, clusters with the highest class proportion are aligned diagonally from class 0. The visualization shows that traditional thresholding causes significant confusion as new classes emerge, with samples scattered across multiple clusters and reduced discovery performance.
}
\end{figure}

To illustrate this, we conduct a toy TTD experiment on MNIST using a two-layer fully connected model. We designate five classes ($0$ to $4$) as known and the remaining five ($5$ to $9$) as unknown.
As shown in Fig.~\ref{fig:mnist_euc}, we first apply thresholding to determine whether a test sample belongs to a new class. The experiment provides a clustering visualization of all 10 classes, along with confusion matrices comparing actual and predicted labels.
Results show that traditional thresholding fail to differentiate among unknown classes. Since the decision boundaries for newly discovered classes are initially unclear, previously identified new classes are often misclassified as different novel classes, leading to poor classification performance.
\section{Method}
\label{sec:method}

\subsection{Overview: A Training-Free Framework}

To mitigate the forgetting of known classes, we keep the model parameters fixed when processing new test data.
To enhance prediction and discovery, we propose a Hash Memory (HM) method, which employs a memory buffer to store the feature of newly discovered test data.
HM utilizes Locality-Sensitive Hashing (LSH) to group similar features and leverages global class prototypes for more accurate classification, particularly in uncertain cases. Additionally, we introduce a memory self-correction mechanism to reassess and reassign labels for stored reference features, ensuring higher reliability.
As shown in Fig.~\ref{fig:mnist_ours}, HM enables the model to effectively discover unknown classes while preserving knowledge of known ones.

\subsection{Local-Sensitively Hashing Memory}

In TTD, distinguishing between seen and unseen categories within unknown classes is challenging. The K-Nearest Neighbor (KNN) classifier~\cite{liao2002use,wang2019research} addresses this by leveraging similarity to previously encountered novel classes, preventing redundant rediscovery while identifying truly new ones through low-density or distant samples in feature space, making it effective for TTD. To achieve this, we construct memory buffers to retain past test data, a proven strategy in TTA~\cite{hong2023mecta} and continual learning~\cite{lopez2017gradient, lyu2023measuring, sun2022exploring, fan2025comprehensive}. To reduce the computational cost of exhaustive comparisons, we integrate an LSH-based method that clusters similar samples into buckets and assigns hash values, enabling efficient retrieval and precise comparisons.

\noindent
\textbf{Hashing memory via feature norm and direction}.
Existing research highlights significant feature norm differences between known and unknown classes, with new class data often exhibiting higher uncertainty and smaller representation norms~\cite{dhamija2018reducing,park2023understanding}. Directional differences are also used to assess sample uncertainty.
Given a test sample $x$, we consider both the scale and direction of feature vectors, constructing the following hashing function:
\begin{equation}
    h(x) = \left[\lfloor\kappa\|f(x)\|\rfloor, \mathbf{1}(f(x)^\top \mathbf{r}_1), \cdots, \mathbf{1}(f(x)^\top \mathbf{r}_n) \right],
    \label{eq:hash}
\end{equation}
where the first term represents the feature scale, discretized by flooring to reduce variations.
The following $n$ terms represent the direction relationship between the feature and $n$ random directions $\{\mathbf{r}_1,\cdots,\mathbf{r}_n\}$.
$\textbf{1}(\cdot)$ is the sign function.
Test samples meeting both scale and direction conditions are assigned to the same bucket, ensuring that their hash values reflect both aspects for efficient retrieval.

\begin{figure*}[t]
\centering

\begin{subfigure}{.5\linewidth}
    \centering
    \includegraphics[width=1\linewidth]{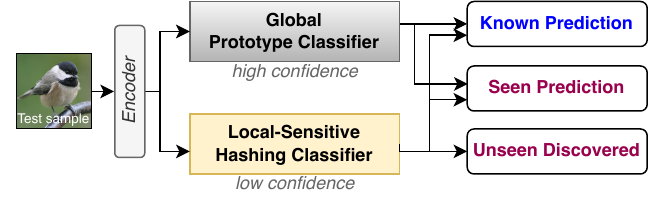}
    \vspace{-15px}
    \caption{Real-time prediction and discovery of our method for TTD.}
    \label{fig:overview}
\end{subfigure}
\hfill
\begin{subfigure}{.45\linewidth}
    \centering
    \includegraphics[width=1\linewidth]{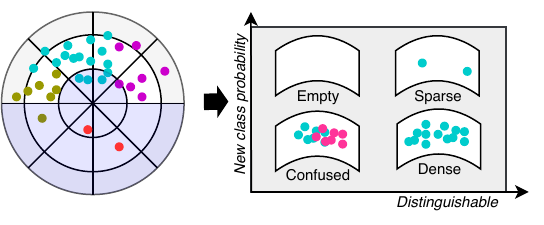}
    \vspace{-15px}
    \caption{Visualization of hash memory in angular space.}
    \label{L2P}
\end{subfigure}
\centering
\begin{subfigure}{\linewidth}
    \centering
    \includegraphics[width=1\linewidth]{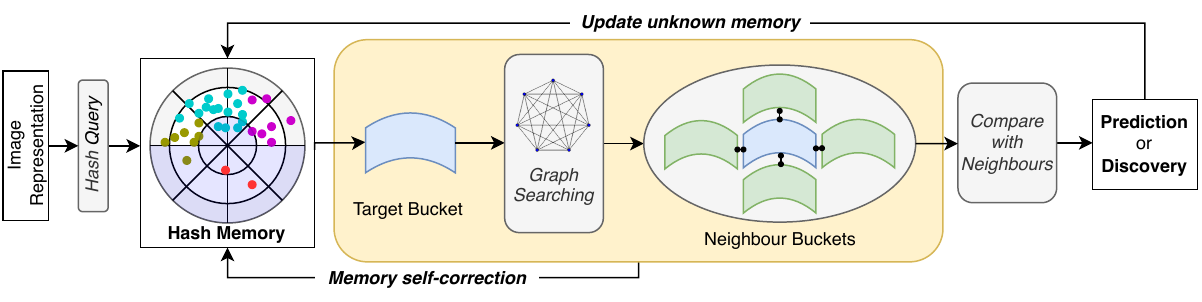}    
    \vspace{-15px}
    \caption{LSH-based classifier and memory self-correction mechanism.}
    \label{DP}
\end{subfigure}
\vspace{-20px}
\caption{Schema of the Proposed Method:
(a) Global-to-local classification: Samples with high confidence are classified using the prototype classifier; otherwise, the LSH-based classifier is used.
(b) Constructed hash memory: Samples are stored in different buckets based on hashed feature norm and direction. Sparse buckets are more likely to contain novel samples.
(c) LSH-based classifier: A test sample is first hashed to locate its target bucket. A graph-based neighbor search then explores adjacent buckets for a more robust prediction. If no similar samples are found, the sample is identified as a new class; otherwise, it is assigned the most relevant label from its neighbors.}
\label{fig:short}
\end{figure*}

\noindent
\textbf{Constructing memory buffer when testing}.
The buffer has a fixed capacity, storing a maximum of the features of $K$ samples per class. 
Two separate memory buffers are maintained: one for known classes sampled from the training set denoted as $\mc{M}_\text{known}=\{\mc{M}_1\cup\mc{M}_2\cup\cdots\cup\mc{M}_{|\mc{Y}_\text{known}|}\}$, and another for unknown classes collected during the testing phase, denoted as $\mc{M}_\text{unknown}=\{\mc{M}_{\#1}\cup\mc{M}_{\#2}\cup\cdots\cup\mc{M}_{\#|\mc{Y}_\text{unknown}|}\}$.
The memory buffer for class $c$ is denoted as,  
\begin{equation}
    \mc{M}_c = \{(h(x_1), f(x_1)), (h(x_2), (x_2)),\cdots\}.
\end{equation}
The features in the memory can be assigned to different buckets according to their hash value.
Given a test sample $x$, the query process can be represented by
\begin{equation}
    \mc{B}(x) = \{(f(x'),y')~|~ h(x')=h(x)\},
\end{equation}
where $y'$ is the corresponding pseudo-label of sample $x'$.
For known classes, the memory is not updated. When a new class is discovered, a new buffer is created for that class, and the memory is updated as additional samples are recognized as belonging to this newly discovered class. 
Since testing is performed online, we adopt the reservoir sampling strategy~\cite{vitter1985random} for memory updates in our implementation.
However, this memory may contain features with wrong labels.
To solve this problem, in Sec.~\ref{sec:sc}, we present a memory self-correction mechanism.

\subsection{Prediction and Discovery based on LSH}

\noindent
\textbf{LSH-based classifier}.
During testing, a sample’s hash value is computed using Eq.~\eqref{eq:hash}.
Then, the sample is compared with the features in the corresponding local bucket to determine its possible class. 
If the matching bucket is empty or sparse, the test sample is considered a potential new class, and a new label is assigned to the sample. The sample's feature and its hash value are then stored in the memory buffer.
This prediction and discovery process can be formulated as follows:
\begin{equation}
y=
\begin{cases}
    y_\text{vote}, & \text{if~} \mc{B}(x) \neq \emptyset, \\
    y_\text{new}, & \text{otherwise},  
    \end{cases}
    \label{eq:lsh}
\end{equation}
where $y_\text{vote} = \mathop{\arg\max}\nolimits_{y\in\mc{Y}_\text{test}} \sum\nolimits_{(f(x'), y') \in \mc{B}(x)} \mathbb{I}(y' = y)$.
If multiple classes share the most samples in the bucket, the one with the smallest average distance is chosen as the voting class.
A newly discovered class can be assigned a non-semantic symbol such as ``$\#C_k$'' or labeled manually, corresponding to the test data point $x$.
We compare generated and annotated labeling in our experiments.

\noindent\textbf{Graph neighbor searching}.
Relying on a single bucket in LSH can result in inaccurate classifications, as it may overlook the data’s underlying structure and fail to account for noise or outliers.
Neighboring buckets provide a broader context, improving generalization, especially in high-dimensional or large-scale datasets.
We propose a graph-based neighbor search strategy, where each bucket is represented as a node, and edges capture directional relationships between bucket representations, defined by the mean feature of internal samples. 
Norms are excluded from graph relationships since they are inherently encoded in the hash values.
The graph dynamically adapts as bucket contents evolve over time. For each target bucket, we select the top-$k$ most Euclidean nearest neighbors to enhance prediction and discovery.
Let $\mc{B}'_1(x), \cdots, \mc{B}'_k(x)$ be the most related neighbor buckets, the joint bucket is denoted as
\begin{equation}
    \hat{\mc{B}}(x) = \mc{B}(x)\cup\mc{B}'_1(x)\cup\cdots\cup\mc{B}'_k(x).
\end{equation}
Then the LSH-based classifier in Eq.~\eqref{eq:lsh} can be applied to the joint bucket.

\noindent\textbf{Global-to-local classifier}.
Although the LSH-based classifier enables finer-grained comparisons when distinguishing between seen and unseen category samples, enhancing accuracy for unknown categories, its overly precise local comparisons can make it challenging for the model to differentiate known categories.
To mitigate this, we first employ a global prototype classifier and then apply the LSH-based classifier according to prediction confidence.
For test samples with high confidence from the prototype classifier (such as using cosine similarity), we directly output the prototype classification result, indicating that the sample is more likely to belong to a known or a seen class. 
For test samples with low confidence, we opt to use the LSH-based classifier (Eq.~\eqref{eq:lsh}) to more precisely determine their possible identity, whether they belong to a new class. In our design, the prototype classifier provides a global classification result, suitable for class recognition with some degree of separability, while LSH offers more refined local sample comparisons, making it better suited for distinguishing hard-to-classify samples.
Although the model is fixed, we update the prototype and the memory buffer for discovered classes.
The prototype is updated by exponential moving average (EMA) with the factor $\alpha$ for a seen class or the feature of a newly discovered class:
\begin{equation}
    \mu_c = \begin{cases}
            \alpha * \mu_c + (1-\alpha) f(x), &\text{if~} c\in\mc{Y}_\text{seen},\\
            f(x), &\text{if~} c\notin\mc{Y}_\text{test}.
    \end{cases}
    \label{eq:prototype}
\end{equation}
Note that we do not update the known prototype to avoid forgetting.
This is similar to some few-shot learning methods~\cite{hajimiri2023strong,sakai2025surprisingly} that only update the prototype classifier, because small changes may lead to catastrophic forgetting.

\subsection{Memory Self-Correction Mechanism}

\label{sec:sc}

The quality of the constructed hash memory is crucial, as it directly impacts the performance of the LSH-based classifier.  
However, the labels assigned to stored features are derived from the classifier, i.e., pseudo-labels, which may not always be reliable, potentially leading to mislabeled samples in the memory.  
To enhance knowledge accumulation and mitigate this issue, we propose a memory self-correction (SC) mechanism, an error correction method designed to refine the hash memory.

The goal of the SC mechanism is to reassign proper labels to stored features in the memory buffer without changing themselves.
This approach is only applicable to unknown classes.
For computational efficiency, we randomly select part (less than 10\%) of memory from each class. 
These selected samples are not used for discovering new classes.
Instead, using the LSH-based classification method, we search for their neighbors within existing classes and assign the appropriate labels.
During this process, samples that are relabeled to known classes or belong to classes that have reached their capacity are discarded. The memory buffer is then replenished with new samples in subsequent test phases.
Specifically, for a memory pair $(f(x), y)\in\mc{M}_\text{unknown}$, its label is reassigned to  
\begin{equation}
    y(x) = \mathop{\arg\max}\limits_{y\in\mc{Y}} \sum\nolimits_{(f(x'), y') \in \hat{\mc{B}}(x)} \mathbb{I}(y' = y),
\end{equation}
which is similar to the LSH-based classification.
If $y(x)\in\mc{Y}_\text{known}$, the pair $(f(x), y)$ will be discarded from the buffer.

\subsection{The Algorithm}

As shown in the Algorithm~\ref{alg:hm}, our method is training free, and built upon three core discovery mechanisms. First, we construct LSH based on the feature differences in norms and directions, so that samples with similar characteristics in the feature space can have the same hash value. This allows the size of the hash bucket to represent the local spatial relationships of the features.
Building on this, we propose using global class prototypes for global class determination. In cases of high uncertainty, hash memory helps with more precise classification. Moreover, the samples stored in the hash memory may initially have a certain probability of errors when discovering new classes. To address this, we introduce a memory self-correction method to reassign labels to the samples in the hash memory.

\begin{figure}[t]
\vspace{-10px}
\begin{minipage}{\linewidth}
\begin{algorithm}[H]
   \caption{Hashing Memory for TTD}
   \label{alg:hm}
\begin{algorithmic}[1]
   \STATE {\bfseries Input:} test sample $x$, {boundary $\epsilon$, prototypes $\mu_{c\in\mc{Y}_\text{test}}$}
   \STATE Compute hash value $h(x)$ using Eq.~\eqref{eq:hash}
   \STATE Search target bucket $\mc{B}(x)$ via the hash value
   \STATE Graph searching for neighboring joint bucket $\hat{\mc{B}}(x)$
   \STATE Compute confidence $u$ via prototype comparison
   \IF{$u>\epsilon$}
   \STATE $y = \arg\max_{c\in\mc{Y}_\text{test}} \frac{f(x)^\top\mu_c}{\|f(x)\|\|\mu_c\|}$ 
   \ELSIF{$\hat{\mc{B}}(x) \neq \emptyset$}
   \STATE $y = \mathop{\arg\max}\nolimits_{y\in\mc{Y}_\text{test}} \sum\nolimits_{(f(x'), y') \in \mc{B}(x)} \mathbb{I}(y' = y)$ 
   \ELSE
   \STATE $y = y_\text{new}$
   {\hfill \textit{\color{teal}\# Novel class discovered}}
   \ENDIF
   
   \STATE Update prototypes via Eq.~\eqref{eq:prototype}
   \STATE Update and self-correct memory buffer
   \STATE {\bfseries Output:} prediction $y$
\end{algorithmic}
\end{algorithm}
\end{minipage}
\end{figure}

\section{Experiment}
\label{sec:ex}

\begin{table*}[t]
\centering
\caption{Major comparisons on CIFAR100D, CUB-200D, and Tiny-ImageNetD. Real-time evaluation reflects the accumulated performance across all test batches, while post evaluation reassesses all test samples after updating the memory and prototypes.}
\vspace{-10px}
\resizebox{\textwidth}{!}{
\begin{tabular}{cl|rrrrr|rrrrrr}
\toprule
\multirow{2}{*}{} &  \multirow{2}{*}{\textbf{Method}} &  \multicolumn{5}{c|}{\textbf{Real-time Evaluation}} &
  \multicolumn{6}{c}{\textbf{Post Evaluation}} \\
\cmidrule{3-13}
  &           & \multicolumn{1}{c}{KA$\uparrow$} & \multicolumn{1}{c}{TA$\uparrow$} & \multicolumn{1}{c}{TE$\downarrow$} & \multicolumn{1}{c}{CA$\uparrow$} & \multicolumn{1}{c|}{CE$\downarrow$} & \multicolumn{1}{c}{KA$\uparrow$} & \multicolumn{1}{c}{TA$\uparrow$} & \multicolumn{1}{c}{TE$\downarrow$} & \multicolumn{1}{c}{CA$\uparrow$} & \multicolumn{1}{c}{CE$\downarrow$} & \multicolumn{1}{c}{KF$\downarrow$} \\
\midrule
\multirow{8}{*}{\rotatebox[origin=c]{90}{\textbf{CIFAR100D}}}                                        & Euclidean & 76.46±0.98 &   17.21±1.33 &  0.52±0.04 &  36.91±3.26 &  2.07±0.41 &  76.62±1.85 &  34.60±2.02 &  1.10±0.04 &  18.70±1.24 &  1.42±0.05 &  -6.45±1.78 \\
        & Cosine & 78.96±1.07 &  \textbf{22.46±1.00} &  0.77±0.01 &  40.85±2.52 &  1.67±0.23 &  78.77±1.87 &  30.60±1.63 &  1.20±0.01 &  29.42±1.46 &  1.62±0.02 &  -5.43±1.82 \\
        & Magnitude  & 75.02±1.12 &  18.00±0.98 &  0.57±0.03 &  36.39±4.43 &  1.76±0.37 &  72.22±1.99 &  35.37±1.75 &  1.27±0.03 &  18.77±1.62 &  \textbf{1.37±0.04} &  -10.85±1.90 \\
        & Entropy   & 77.83±1.07 &  19.54±1.00 &  0.60±0.02 &  42.15±3.85 &  1.57±0.40 &  76.35±2.20 &  \textbf{35.60±2.00} &  1.36±0.02 &  26.79±1.70 &  1.55±0.03 &  -6.72±2.13 \\

&  L2P~\cite{wang2022learning} &  59.93±2.15 &  8.57±2.49 &  0.60±0.06 &  43.10±4.30 &  1.85±0.18 &  50.53±7.25 &  9.60±1.50 &  0.77±0.12 &  27.39±2.11 &  1.37±0.24 &  -27.85±7.21 \\
& DP~\cite{wang2022dualprompt}        &66.09±1.01       & 8.80±1.69 & 0.53±0.08 & 48.34±6.78 & 1.63±0.30 & 56.19±2.00 & 8.68±2.06 & 0.70±0.08 & 28.93±2.25 & 1.34±0.05 & -29.06±1.99    \\
& GMP~\cite{cendra2025promptccd}       & 72.77±1.20     & 7.37±0.88 & \textbf{0.51±0.05} & 42.26±4.54 & 1.80±0.24 &  67.21±2.53 & 13.04±2.59 & 1.18±0.06 & 27.10±1.77 & 1.55±0.08 & -17.69±2.53   \\
   & Ours      & \textbf{79.17±0.13} &  21.13±0.62 &  0.67±0.02 &  \textbf{56.37±1.42} &  \textbf{1.23±0.08} & \textbf{80.73±1.59} &  31.03±1.24 &  \textbf{1.07±0.02} &  \textbf{34.81±1.22} &  1.50±0.02 &  \textbf{-3.41±1.49}\\
\midrule
\multirow{8}{*}{\rotatebox[origin=c]{90}{\textbf{CUB200D}}}        & Euclidean & 66.09±1.20 &  43.60±2.08 &  0.40±0.06 &  44.05±4.96 &  1.46±0.40 &  65.52±3.68 &  49.90±1.33 &  0.81±0.00 &  6.64±0.67 &  0.70±0.00 &  \textbf{-1.61±3.55}\\
& Cosine & 52.06±1.04 &  37.87±1.90 &  0.43±0.04 &  58.63±3.55 &  1.04±0.28 &  41.86±2.27 &  25.77±0.96 &  0.66±0.01 &  29.84±1.64 &  1.98±0.00 &  -25.42±2.17  \\
& Magnitude  & 55.05±2.20 &  59.45±2.42 &  0.65±0.06 &  47.96±3.12 &  1.42±0.34 &  59.24±3.33 &  47.03±1.21 &  1.68±0.00 &  9.34±0.85 &  1.12±0.01 &  -7.89±3.20\\
& Entropy   & 62.13±1.60 &  60.34±2.22 &  0.69±0.04 &  47.86±2.71 &  1.41±0.30 &  59.71±2.96 &  48.69±1.54 &  1.78±0.01 &  12.02±1.10 &  1.18±0.02 &  -7.42±2.83\\
& L2P~\cite{wang2022learning}       & 46.22±1.53 &  9.01±0.87 &  0.44±0.02 &  \textbf{55.37±7.79} &  \textbf{0.97±0.25} &  31.97±3.35 &  4.75±0.73 &  \textbf{0.51±0.03} &  24.48±1.65 &  \textbf{0.62±0.06} &  -42.29±3.14 \\
& DP~\cite{wang2022dualprompt}         & 53.69±1.24 &  43.68±2.20 &  0.40±0.06 &  45.68±5.88 &  1.50±0.36 &  63.37±3.27 &  45.94±0.91 &  1.69±0.02 &  7.92±1.10 &  1.10±0.03 &  -5.85±3.11 \\
& GMP~\cite{cendra2025promptccd}       &  62.97±1.33 &  46.44±1.87 &  0.59±0.03 &  47.99±4.34 &  1.49±0.13 &  58.11±3.00 &  48.02±1.20 &  1.53±0.01 &  10.31±1.45 &  0.90±0.00 &  -5.46±2.77\\
 & Ours      & \textbf{66.20±0.55} &  \textbf{58.30±2.37} &  \textbf{0.35±0.02} &  43.33±6.10 &  1.92±0.83 &  \textbf{64.42±0.65} &  \textbf{65.28±1.78} &  1.02±0.03 &  \textbf{37.25±4.90} &  1.24±0.22 &  -4.07±0.47 \\
\midrule
\multirow{8}{*}{\rotatebox[origin=c]{90}{\textbf{Tiny-ImageNetD}}} & Euclidean & 57.53±1.80 &  11.35±1.56 &  0.48±0.03 &  63.31±3.55 &  0.66±0.12 &  52.36±3.10 &  6.48±1.40 &  \textbf{0.37±0.02} &  13.81±2.11 &  \textbf{0.44±0.02} &  -22.90±3.08 \\
& Cosine & 62.85±1.37 &  10.12±0.40 &  0.50±0.00 &  70.97±5.27 &  \textbf{0.57±0.03} &  56.87±2.98 &  10.00±1.10 &  0.84±0.00 &  32.80±1.25 &  1.28±0.02 &  -19.22±2.98  \\
& Magnitude  & 63.79±2.00 &  10.08±1.10 &  0.48±0.02 &  53.93±2.69 &  1.17±0.08 &  57.80±2.21 &  11.97±1.11 &  0.91±0.02 &  17.19±1.68 &  0.87±0.03 &  -17.51±2.19 \\
& Entropy   & 63.14±1.54 &  15.52±0.87 &  0.48±0.00 &  53.56±3.01 &  1.19±0.06 &  61.66±3.02 &  13.77±1.41 &  1.07±0.01 &  15.32±1.40 &  0.95±0.03 &  -13.65±3.00 \\
& L2P~\cite{wang2022learning}       & 46.25±1.41 &  7.79±2.92 &  0.51±0.03 &  53.55±6.47 &  1.33±0.23 &  29.50±3.77 &  10.38±2.42 &  0.89±0.06 &  23.28±0.79 &  1.37±0.06 &  -47.97±3.77 \\
& DP~\cite{wang2022dualprompt}         &46.51±0.58 &  6.41±0.93 &  0.51±0.03 &  58.27±6.10 &  1.15±0.21 &  28.53±3.33 &  9.63±2.10 &  0.85±0.02 &  26.80±1.38 &  1.33±0.02 &  -47.57±3.32\\
& GMP~\cite{cendra2025promptccd}       &62.47±1.40 &  6.25±1.72 &  \textbf{0.45±0.02} &  58.02±4.29 &  1.08±0.14 &  63.95±2.04 &  15.64±2.63 &  1.30±0.03 &  26.31±2.33 &  1.54±0.03 &  -16.86±2.04\\
& Ours      & \textbf{75.31±1.31} &  \textbf{16.04±0.76} &  0.51±0.00 &  \textbf{73.81±2.67} &  0.61±0.04 &  \textbf{74.94±2.20} &  \textbf{16.23±1.24 }&  0.81±0.00 &  \textbf{37.43±1.30} &  1.21±0.02 &  \textbf{-1.15±2.18} \\
\bottomrule
\end{tabular}
}
\label{tab:major}
\end{table*}

\subsection{Experimental Details}
\label{sec:experimental details}

\noindent\textbf{Dataset construction}.
We conduct our experiments based on three benchmark datasets, namely CIFAR100 (C100)\cite{cifar100krizhevsky2009learning}, Caltech-UCSD Birds-200-2011 (CUB)\cite{cubwah2011caltech} and Tiny ImageNet\cite{tinyle2015tiny}. 
All these datasets are split into known and unknown classes (7:3).
The model is trained on the known training set, and tested on the mixture of known and unknown test sets.
Because the dataset is used for discovery, we name the three transformed datasets CIFAR100D, CUB-200D, and Tiny-ImageNetD.
More details of the dataset construction can be seen in \app{Appendix B}.

\noindent\textbf{Implementation details}.
In our implementation, we build our method on the prompt-based method L2P~\cite{wang2022learning}, which employs a ViT-B/16 backbone pretrained with DINO. 
To improve the representation, when we fine-tune the retained model on the known classes, we follow objectosphere~\cite{dhamija2018reducing} to reduce the norm in the loss function.

\noindent\textbf{Evaluation metrics}.
First, we provide \textbf{real-time evaluation}, including the final accumulated value. Then, since traditional NCD methods rely on post evaluation, we also present \textbf{post evaluation}, which revalidate all test samples collectively.
The evaluations are split into two parts: one for known classes and one for unknown classes. 
For known classes, we use traditional accuracy (\textit{Known Accuracy}, \textbf{KA}) and accuracy forgetting (\textit{Known Forgetting}, \textbf{KF}). 
For unknown classes, since the predicted label space $\mc{Y}^\text{GT}_\text{seen}$ does not match the cluster label space $\mc{Y}_\text{seen}$, we propose agreement metrics to assess effectiveness, where $\mc{Y}^\text{GT}_\text{seen}$ represent the true label space and $\mc{Y}_\text{seen}$ the predicted label space.
In the test set $\mc{D}^\text{test}$, a sample $x$ has a true label $y\in\mc{Y}^\text{GT}_\text{seen}$ and a predicted cluster label $p(x)\in\mc{Y}_\text{seen}$. For simplicity, we define the subset of $\mc{D}^\text{test}$ with true label $c$ as $\mc{D}^\text{test}_c$, and the cluster with predicted label $q$ as $\mc{C}^\text{test}_q$.

\noindent
(1) \textit{True-label Agreement ratio} (\textbf{TA}). This metric measures the maximum proportion of samples from a given true class that are predicted as the same class:
\vspace{-5px}
\begin{equation*}
    \text{TA} = \mbb{E}_{c \in \mc{Y}^\text{GT}_\text{seen}} \frac{1}{|\mc{D}^\text{test}_c|}\max_{q\in\mc{Y}_\text{seen}}\left(\sum\nolimits_{x\in\mc{D}^\text{test}_c}\mb{1}(p(x)=q)\right).
\vspace{-5px}
\end{equation*}
\noindent
(2) \textit{True-label Entropy} (\textbf{TE}). This metric measures the average entropy $H(\cdot)$ of the predicted labels for samples with that true class:
\vspace{-5px}
\begin{equation*}
    \text{TE} =\mbb{E}_{c \in \mc{Y}^\text{GT}_\text{seen}}H(\{p(x)|x\in\mc{D}^\text{test}_c\}).
\vspace{-5px}
\end{equation*}
\noindent
(3) \textit{Cluster Agreement ratio} (\textbf{CA}). This metric measures the maximum proportion of samples from a given predicted cluster that are with the same true label:
\vspace{-5px}
\begin{equation*}
    \text{CA} = \mbb{E}_{q \in \mc{Y}_\text{seen}} \frac{1}{|\mc{C}^\text{test}_q|}\max_{c\in\mc{Y}^\text{GT}_\text{seen}}\left(\sum\nolimits_{(x,y)\in\mc{C}^\text{test}_q}\mb{1}(y=c)\right).
\vspace{-5px}
\end{equation*}
\noindent
(4) \textit{Cluster Entropy} (\textbf{CE}). This metric measures the average entropy of the samples that predicted the true class contained in clusters:
\vspace{-5px}
\begin{equation*}
    \text{CE} =\mbb{E}_{q \in \mc{Y}_\text{seen}}H(\{y|(x,y)\in\mc{C}^\text{test}_q\}).
\vspace{-5px}
\end{equation*}
For more details of the metrics, see \app{Appendix C}.

\subsection{Major Comparisons}

In this paper, we first compare our methods with some thresholding-based training-free methods, including Euclidean distance between prototypes, cosine similarity between prototypes, feature magnitude, and logit entropy. When these methods exceed the threshold, they will be considered to have discovered a new class. The threshold tuning for these methods is provided in \app{Appendix D}.
We also compare with some training-required methods including L2P~\cite{wang2022learning}, DP~\cite{wang2022dualprompt}, and GMP~\cite{cendra2025promptccd}, these methods update prompts like TTA and CL methods.

The results are shown in Table~\ref{tab:major}, we have some major observations.
First, real-time and post evaluations exhibit notable differences, with post evaluation showing higher TA but lower CA. This suggests that in NCD with large sample sizes, post evaluation better distinguishes class differences, grouping similar samples into clusters. However, clusters often contain multiple classes, indicating confusion in new class discovery.
Second, compared to traditional threshold-based training-free methods, training-based methods suffer from performance degradation. Immediate model updates upon discovering new classes lead to lower TA and CA, along with more severe catastrophic forgetting.
Finally, our method outperforms others across all three datasets, demonstrating that by leveraging fine-grained sample-level comparisons and a memory self-correction mechanism, our approach enhances adaptability in novel class discovery.

\subsection{Analysis on TTD}

\noindent\textbf{Different number of discoverable categories}.
In our experiments, we set an upper limit on discoverable classes, though real-world scenarios may involve a much larger or even infinite number. We tested three cases: (1) reducing unknown classes while keeping known classes fixed, (2) increasing known classes while keeping unknown classes unchanged, and (3) exceeding the true number of unknown classes. Table~\ref{tab:old+new} presents the results.
We find that more discoverable classes generally improve TA and CA, though the effect depends on the number of known classes. When the discoverable classes far exceed the true number, real-time evaluation shows TA and CA gains, but post evaluation sees a TA drop and more severe forgetting. These findings highlight the importance of appropriately setting the number of discoverable classes for optimal TTD performance.

\begin{table}[t]
\centering
\caption{Comparisons of different discoverable class numbers.}
\vspace{-10px}
\resizebox{\linewidth}{!}{
\begin{tabular}{l|cccc|ccccc}
\toprule
Known + & \multicolumn{4}{c|}{\textbf{Real-time Eval}} & \multicolumn{5}{c}{\textbf{Post Eval}}       \\ \cmidrule{2-10} 
Unknown   & TA       & TE     & CA      & CE     & TA    & TE   & CA    & CE   & KF    \\
\midrule
70+10             & 13.70    & 0.55   & 55.70   & 1.23   & 16.00 & 0.73 & 21.09 & 1.08 & -1.33 \\
70+20             & 15.89    & 0.60   & 55.13   & 1.25   & 22.90 & 0.99 & 28.84 & 1.40 & -2.54 \\
70+30             & 21.11    & 0.66   & 56.87   & 1.27   & 31.03 & 1.07 & 34.81 & 1.50 & -3.47 \\
\midrule
80+20             & 11.57    & 0.50   & 64.15   & 1.00   & 20.90 & 0.51 & 31.09 & 0.87 & -0.69 \\
90+10             & 14.92    & 0.45   & 52.98   & 1.22   & 21.50 & 0.38 & 30.61 & 0.57 & -0.44  \\
\midrule
70+30+100           & 17.58                                            & 0.70    & 81.28    & 0.44   & 25.03 & 1.82 & 40.74 & 1.05 & -6.46  \\
70+30+200           & 19.86                                            & 0.76    & 85.60    & 0.33   & 26.87 & 2.17 & 42.63 & 0.84 & -7.79  \\
70+30+$\infty$          & 20.37                                            & 0.84    & 92.94    & 0.16   & 22.63 & 2.86 & 47.09 & 0.46 & -10.69 \\
\bottomrule
\end{tabular}
}
\label{tab:old+new}
\end{table}

\noindent\textbf{Comparisons on NCD metrics}.
Traditional NCD methods use post-cluster evaluation, we also provide some cluster metrics, including Hungarian Cluster Accuracy (HCA)~\cite{meilua2003comparing}, Adjusted Rand Index (ARI)~\cite{rand1971objective}, Normalized Mutual Information (NMI)~\cite{mcdaid2011normalized} and V-Measure~\cite{rosenberg2007v}.
More details can be seen in \app{Appendix C.3}.
The results are shown in Table~\ref{fig:othermetric}, which indicate that the proposed method still performs well when evaluated using the classic NCD clustering accuracy metric.

\begin{figure}[h]
\centering
\vspace{-10px}
\includegraphics[width=\linewidth]{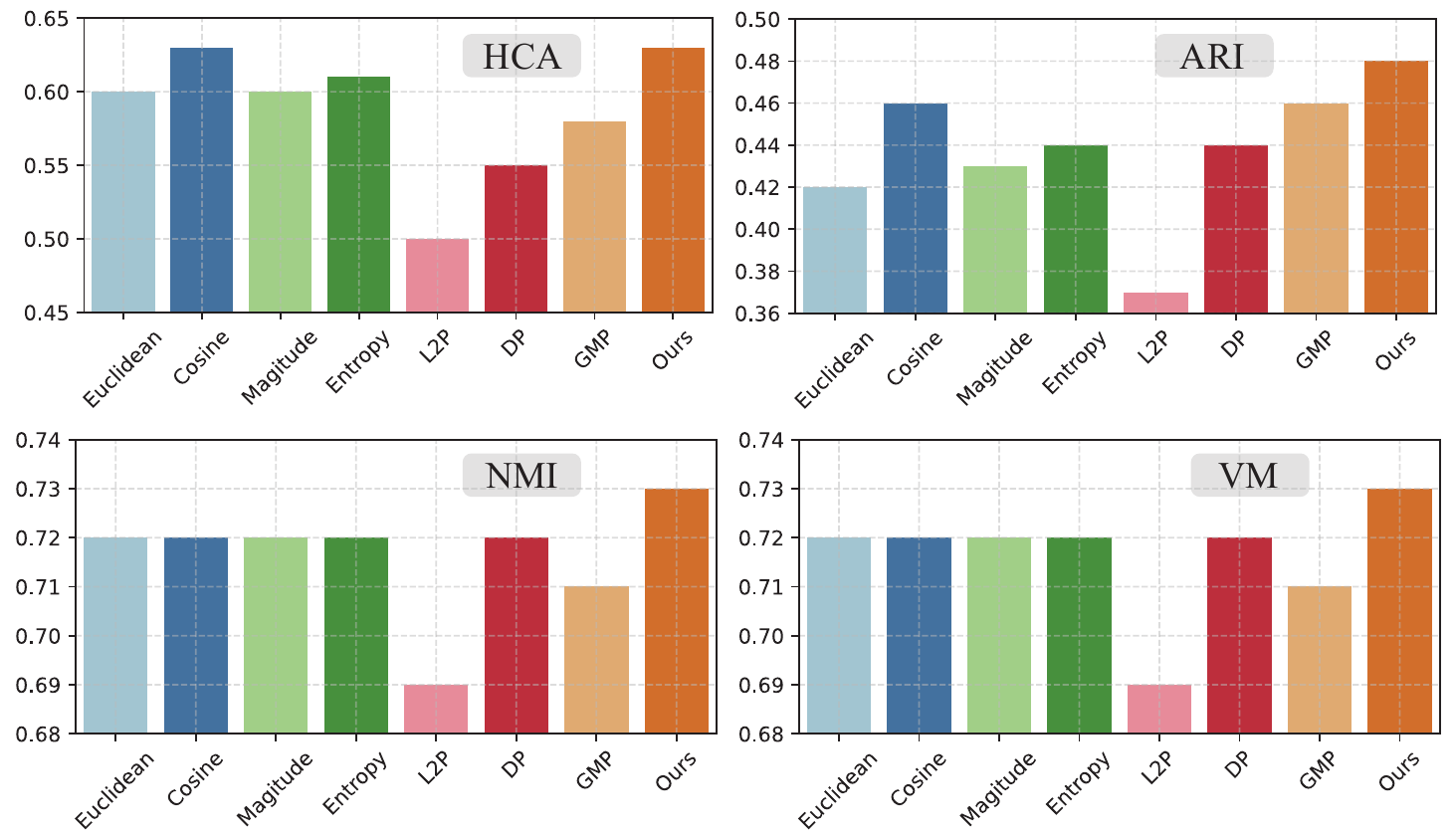}
\vspace{-20px}
\caption{Post evaluation using NCD metrics.}
\label{fig:othermetric}
\end{figure}

\noindent\textbf{Human annotation vs. Auto label assignment}.
In the above experiments, newly discovered classes are automatically assigned non-semantic labels. However, when human annotations are provided for samples potentially belonging to new classes, semantic labels eliminated erroneous discoveries. Table~\ref{tab:Human}(a) presents this experiment, showing that with human annotations, the TA for novel class discovery significantly improved (21.11 vs. 52.10), while CA declined. This suggests that human annotations strongly enhance intra-class consistency but do not improve real-time classification performance.

\begin{table}[t]
\centering
\caption{Comparison across multiple experimental settings, including (a) human annotation vs. automatic assignment, (b) training-required vs. training-free setting, and (c) PCA-based vs. random direction representations.}
\vspace{-10px}
\resizebox{\linewidth}{!}{
\begin{tabular}{c@{\hspace{2pt}}l|cccc|ccccc}
\toprule
&\multirow{2}{*}{\textbf{Experiment}} & \multicolumn{4}{c|}{\textbf{Real-time Eval}} & \multicolumn{5}{c}{\textbf{Post Eval}}       \\ \cmidrule{3-11} 
                  && TA       & TE     & CA      & CE     & TA    & TE   & CA    & CE   & KF    \\
\midrule
(a)&Human annotation     & 52.10 & 0.48 & 42.96 & 1.68 & 48.27 & 1.19 & 49.44 & 1.24 & -5.81 \\ 
(b)&Training-required   & 43.25     & 1.09      & 12.88     & 4.17      & 9.83      & 0.87      & 33.69     & 1.61      & -3.17 \\
(c)&{PCA bases}     & 20.86   & 0.65   & 46.33   & 1.09   & 28.25  & 1.45  & 28.91 & 1.66 & -3.24 \\
\midrule
 & Ours      & 21.11 & 0.66 & 56.87 & 1.27 & 31.03 & 1.07 & 34.81 & 1.50 & -3.47 \\
\bottomrule
\end{tabular}
}
\label{tab:Human}
\end{table}

\noindent\textbf{Training-free vs. training-required TTD}.
In Table~\ref{tab:Human}(b), we explore the impact of allowing prompt training while using HM. The results reveal a sharp decline in CA and a rise in TA after training, suggesting that most new classes are misclassified as known ones. This highlights how model updates in TTD directly affect known class performance, leading to a rapid degradation in overall recognition when new classes remain uncertain.

\subsection{Analysis on Hash Memory}

\noindent\textbf{Different memory sizes}.
Table~\ref{tab:memory} presents the effect of different memory sizes in HM. As memory size increases, TA decreases while CA increases, suggesting that excessive memory may impair the correct rediscovery of new classes. 
This is likely due to the reliance on pseudo-labeling in memory storage, where rapidly storing a large number of new class samples increases the risk of misclassification.

\begin{table}[h]
\vspace{-10px}
\centering
\caption{Comparison of different memory size.}
\vspace{-10px}
\resizebox{\linewidth}{!}{
\begin{tabular}{c|cccc|ccccc}
\toprule
Memory size & \multicolumn{4}{c|}{\textbf{Real-time Eval}} & \multicolumn{5}{c}{\textbf{Post Eval}}       \\ \cmidrule{2-10} 
(per class)       & TA       & TE     & CA      & CE     & TA    & TE   & CA    & CE   & KF    \\
\midrule
0                 & 26.56    & 0.77   & 40.85   & 1.67   & 30.60 & 1.19 & 29.42 & 1.62 & -5.43 \\
10                & 22.33    & 0.70   & 53.18   & 1.41   & 28.47 & 1.12 & 28.02 & 1.54 & -3.71 \\
20                & 21.11    & 0.66   & 56.87   & 1.27   & 31.03 & 1.07 & 34.81 & 1.50 & -3.47 \\
30                & 15.39    & 0.66   & 66.99   & 0.89   & 22.57 & 1.10 & 33.05 & 1.56 & -3.44  \\
\bottomrule
\end{tabular}
}
\label{tab:memory}
\end{table}

\noindent\textbf{PCA direction bases vs. random direction bases}.
In our method, we use randomly assigned reference directions when constructing hash values. In Table~\ref{tab:Human}(c), we test an alternative approach using PCA to obtain reference directions from known classes. The results show that PCA-based directions perform similarly to random directions.

\begin{figure}[t]
\centering
\includegraphics[width=\linewidth]{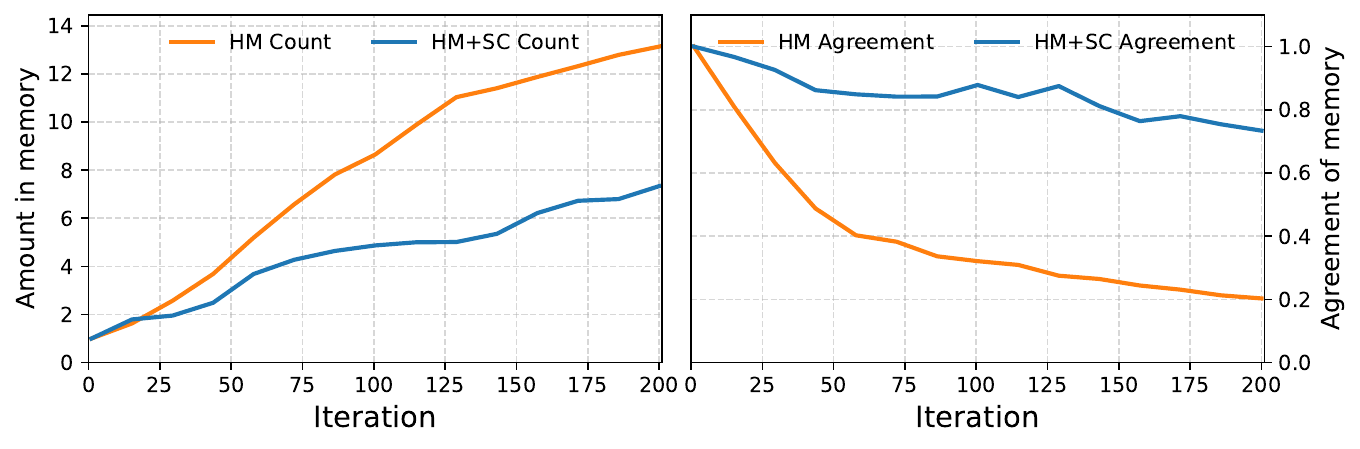}
\vspace{-20px}
\caption{Memory agreement w/ and w/o SC.}
\label{fig:ex_sc}
\end{figure}

\noindent\textbf{Memory agreement w/ and w/o SC}.
In Fig.~\ref{fig:ex_sc}, we compare the impact of using and not using the SC strategy on memory. The left plot shows the number of true novel class samples in the buffer, while the right plot illustrates the consistency of true labels within each buffer class. The results indicate that SC effectively corrects mislabeled samples in the buffer, enhancing memory performance.

\begin{figure}[h]
\centering
\includegraphics[width=\linewidth]{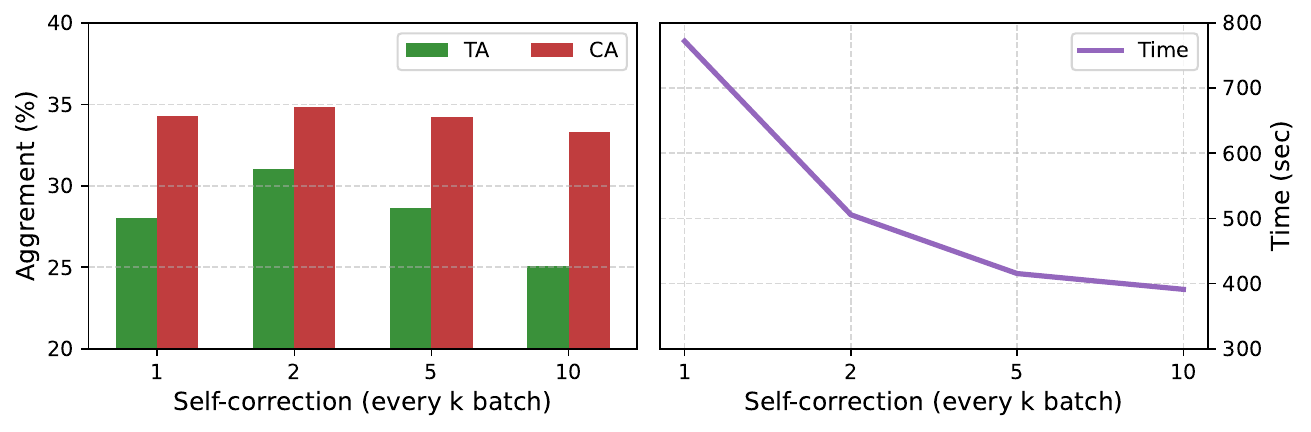}
\vspace{-20px}
\caption{Frequency of memory self-correction.}
\label{fig:sc_frequency}
\end{figure}

\noindent\textbf{Frequency of SC}.
In Fig.~\ref{fig:sc_frequency}, we examine the effect of different SC frequencies. The results show that frequent self-correction increases computation time without significantly improving novel class discovery. Conversely, infrequent self-correction leads to the accumulation of erroneous pseudo-labels, negatively impacting model performance.

\subsection{Analysis on LSH Classifier}

\noindent\textbf{Ablation study}.
Table~\ref{tab:ablation} presents the ablation study results, showing that the proposed HM improves performance over the prototype classifier, particularly in CA. Adding SC further enhances performance, as it increases the reliability of sample labels stored in memory.

\begin{table}[h]
\centering
\caption{Ablation study.}
\vspace{-10px}
\resizebox{\linewidth}{!}{
\begin{tabular}{cc|cccc|ccccc}
\toprule
\multirow{2}{*}{HM}  & \multirow{2}{*}{SC}   & \multicolumn{4}{c|}{\textbf{Real-time Eval}}                                                                         & \multicolumn{5}{c}{\textbf{Post Eval}}                                                                                                           \\ \cmidrule{3-11} 
                     &                       &  TA       & TE     & CA      & CE     & TA    & TE   & CA    & CE   & KF                        \\
\midrule
                                         
                    &                     & 26.56    & 0.77   & 40.85   & 1.67   & 30.60 & 1.19 & 29.42 & 1.62 & -5.43 \\
$\checkmark$     &       & 22.18    & 0.75   & 52.51   & 1.44   & 31.21 & 1.45 & 31.01 & 1.47 & -4.79 \\
$\checkmark$     & $\checkmark$    & 21.11    & 0.66   & 56.87   & 1.27   & 31.03 & 1.07 & 34.81 & 1.50 & -3.47  \\
\bottomrule
\end{tabular}
}
\label{tab:ablation}
\end{table}

\begin{figure}[t]
\centering
\includegraphics[width=\linewidth]{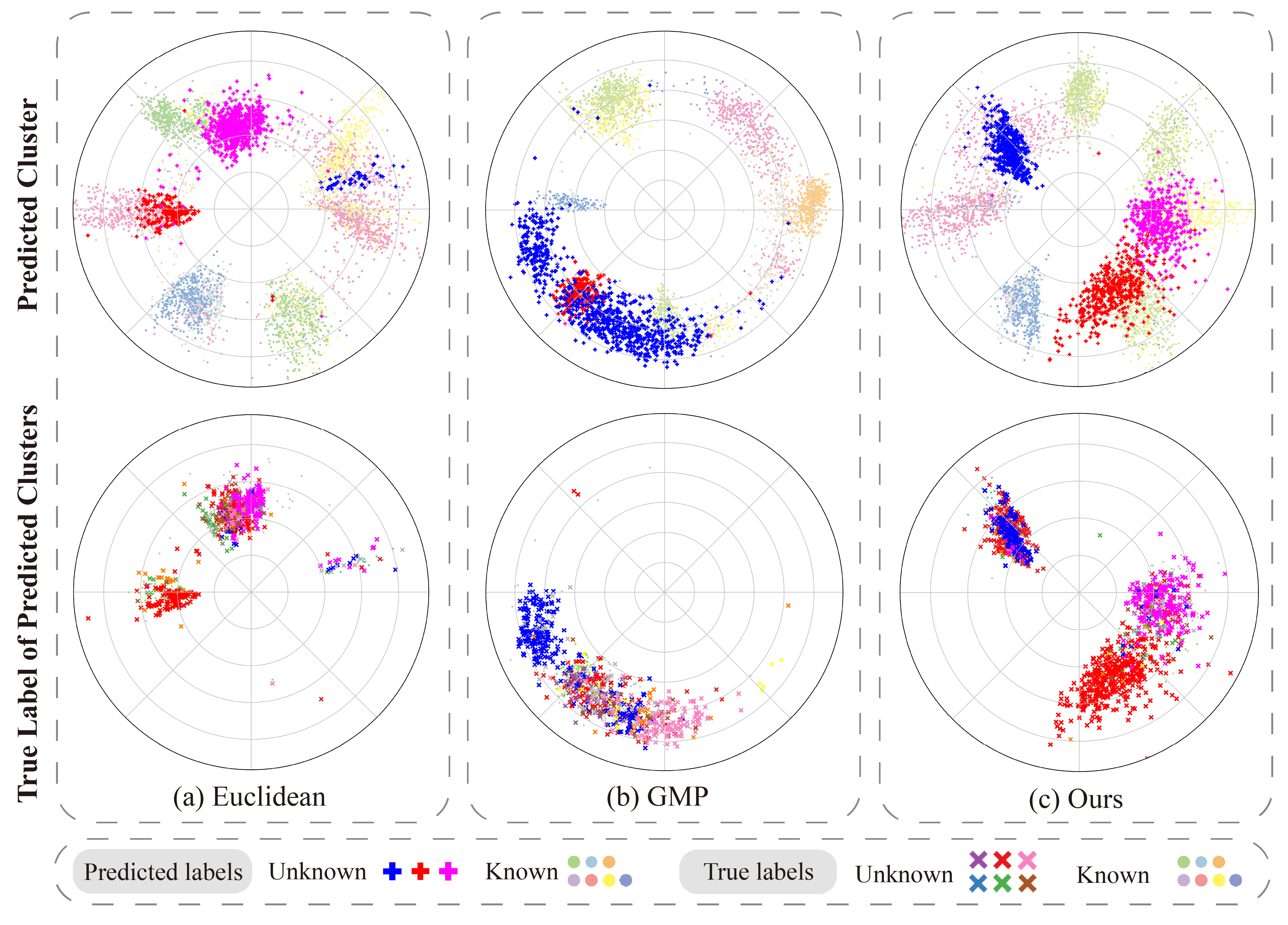}
\vspace{-20px}
\caption{Post t-SNE visualization in angular space on CIFAR100D with 7 known and 3 unknown classes.
The first column is colored by predicted clusters.
In the second column, within each cluster, the true label that appears most frequently is assigned the same color as the cluster  in the first column.}
\label{fig:tsne}
\end{figure}

\noindent\textbf{Post visualization using t-SNE}.
In Fig.~\ref{fig:tsne}, we employ t-SNE~\cite{van2008visualizing} to visualize the true-label distribution of test samples, comparing predicted clusters with their corresponding ground-truth labels. The results demonstrate that our method effectively distinguishes between known and unknown classes, whereas baseline methods often confuse them, compromising both novel class discovery and known class classification. Additionally, our approach successfully groups samples with the same labels into cohesive clusters, a crucial factor for effective novel class discovery.

\begin{figure}[t]
    \centering
    \begin{minipage}[c]{0.45\linewidth}
        \centering
        \includegraphics[width=\linewidth]{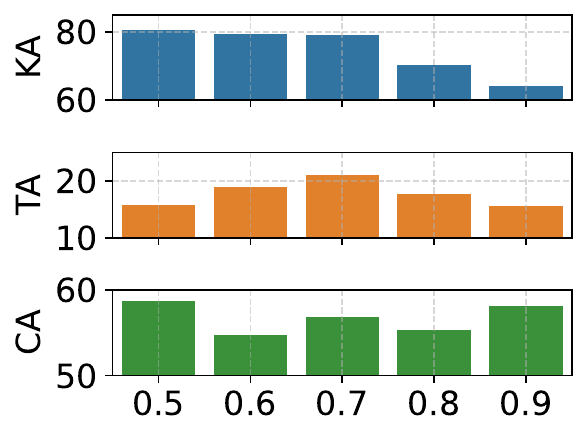}
        \vspace{-20px}
        \caption{Boundary between prototype and LSH classifiers}
        \label{fig:boundary}
    \end{minipage}
    \hspace{2px}
    \begin{minipage}[c]{0.52\linewidth}
        \centering
        \includegraphics[width=\linewidth]{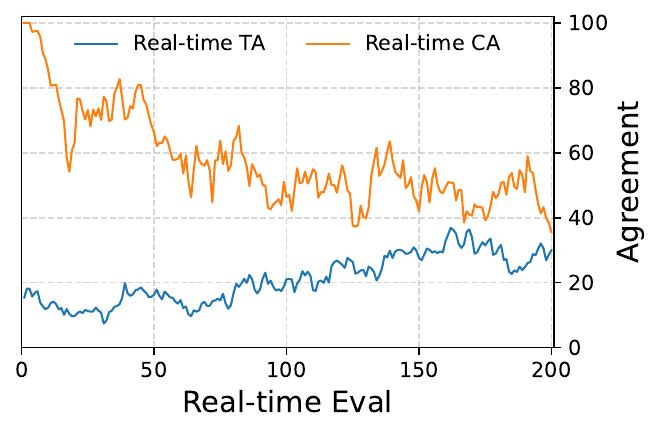}
        \vspace{-20px}
        \caption{Trends of TA and CA across the whole TTD process.}
        \label{fig:trend}
    \end{minipage}%
\end{figure}

\noindent\textbf{Boundary between prototype and LSH classifiers}.
Our method employs a global-to-local classification approach, where confidence determines whether to use the prototype classifier or the LSH-based classifier. 
In Fig.~\ref{fig:boundary}, we analyze the effect of varying this boundary value. The results show that with a higher boundary, \ie favoring the LSH-based classifier, KA drops sharply. Conversely, with a lower boundary, \ie favoring the prototype classifier, novel class discovery becomes unstable.

\noindent\textbf{Trends of TA and CA across the whole TTD}.
In Fig.~\ref{fig:trend}, we illustrate how real-time TA and CA evolve during testing. The results show a continuous decline in TA and a steady increase in CA, aligning with the novel class discovery process. Early on, clusters contain fewer samples, leading to more pseudo-labeling errors. Over time, as more test samples accumulate, TA and CA gradually reach a balance.

\section{Conclusion}

This paper introduces Test-Time Discovery (TTD), a novel paradigm for real-time class adaptation, addressing the limitations of TTT and NCD. 
To tackle key challenges in TTD, we propose a training-free, hash-based memory mechanism that leverages locality-sensitive hashing for efficient sample retrieval and collaborative classification for improved accuracy. Moreover, a self-correction mechanism enhances memory reliability. Experimental results show that our approach outperforms existing methods in test-time scenarios, enabling robust novel category discovery and classification.
Although our method achieves good results in test-time discovery, it has some \textit{limitations}. First, the method relies on dynamically constructing a hash memory during testing, which may be unsuitable for privacy-sensitive scenarios. Second, using feature norm and direction for hashing may not be optimal, and future work could explore more effective hashing strategies.
{
    \small
    \bibliographystyle{ieeenat_fullname}
    \bibliography{main}
}

\appendix

~

~

{\Large\noindent\textbf{Appendix}}

\section{Detailed Related Work}
\label{sec:rel}

\subsection{Test-Time Learning}

\label{sec:rel_ttl}

\textit{Test-Time Training} (TTT) is a machine learning approach where a model continues to update its parameters during the test phase by leveraging auxiliary tasks. These tasks, often self-supervised, are designed to exploit the structure of the input data and improve the model’s adaptability to test-time distribution shifts. 
By incorporating additional training during inference, TTT can adjust specific components, such as feature extractors, to enhance performance on out-of-distribution or domain-shifted data. 
TTT~\cite{sun2020test} trains a model jointly on the task loss and an auxiliary image-rotation prediction task during training, which couples the gradients of the auxiliary task and the task itself.
TTT++~\cite{liu2021ttt++} further enhances performance during test time. The statistics of the source data are calculated from a retained queue of source feature maps. Subsequently, these statistics are employs to align with the target data, thereby regularizing the contrastive loss.
TTT-MAE~\cite{gandelsman2022test} utilized a masked Autoencoder to predict the obscured parts, thus tuning the model at test time.
TTTFlow~\cite{osowiechi2023tttflow} employs Normalizing Flows to detect domain shifts and adapt deep neural networks for improved accuracy on distributionally shifted data.
TTAC++~\cite{su2024revisiting} employs anchored clustering to discover clusters in both source and target domains and matches the target clusters to the source ones for improved adaptation. When source domain information is strictly absent, the method can efficiently infer source domain distributions to achieve source-free test-time training. 
However, this approach typically requires additional computation and well-designed auxiliary objectives.

\textit{Test-Time Adaptation} (TTA) focuses on dynamically adjusting a model during inference without explicit training. This method usually updates lightweight parameters, such as batch normalization statistics, or applies simple strategies like entropy minimization to adapt the model to new test data distributions. 
TENT~\cite{wangtent} conducts its adaptation on the batch normalization layers using the conditional entropy loss of the predictions. 
SHOT~\cite{liang2020we} only needs a well-trained source model and achieves domain adaptation without accessing potentially private source data by leveraging information maximization and self-supervised pseudo-labeling techniques.
LAME~\cite{boudiaf2022parameter} adjusts the classifier's output rather than its internal parameters, utilizing Laplacian regularization to encourage consistent latent assignments for neighboring points in the feature space.
AdaContrast~\cite{chen2022contrastive} refines pseudo labels through soft voting among nearest neighbors in the target feature space and utilizes contrastive learning to exploit pairwise information among target samples.
RoTTA~\cite{yuan2023robust} employs a robust batch normalization scheme to estimate normalization statistics and utilizes a memory bank to capture a snapshot of the test distribution through category-balanced sampling that considers timeliness and uncertainty. Moreover, RoTTA develops a timeliness-aware reweighting strategy along with a teacher-student model to stabilize the training process.
TTA aims to improve the model's robustness and accuracy while keeping computational overhead minimal, making it particularly suitable for real-time applications or scenarios with mild distribution shifts.

In summary, TTT and TTA tasks focus solely on handling changes in data distribution and do not consider scenarios where new classes are encountered during testing. In open-world testing environments, models are likely to encounter new classes at any time, and the ability to autonomously discover these new classes would enhance the model’s adaptability to a broader range of unknowns.

\subsection{Novel Category Discovery}

Novel Category Discovery (NCD) focuses on enabling models to autonomously detect and adapt to new, previously unseen classes during the inference phase, particularly in open-world environments. NCD allows models to identify when a new category appears and respond appropriately, often without labeled data for the new class. The challenge lies in ensuring that models can effectively detect and represent novel categories while maintaining the integrity of previously learned knowledge.
DTC~\cite{han2019learning} leverages prior knowledge of related but distinct image classes, reduces ambiguity in clustering, and enhances the quality of newly discovered classes.
Han et al.~\cite{han2020automatically} introduced a novel approach that combines self-supervised learning, ranking statistics, and joint optimization to automatically discover and learn new visual categories in an image collection.
Zhao et al.~\cite{zhao2021novel} proposed using dual ranking statistics and mutual knowledge distillation, generating pseudo labels through a dynamically constructed local part dictionary, and allowing information exchange between the two branches to encourage agreement on new category discovery, thus enabling the model to leverage the benefits of both global and local features.
GCD~\cite{vaze2022generalized} leverages contrastively trained Vision Transformers to directly assign labels through clustering, avoiding overfitting to labeled classes.
PromptCAL~\cite{zhang2023promptcal} performs contrastive affinity learning with auxiliary prompts and attribute semantic self-supervised learning to enhance visual feature discrimination for generalized novel category discovery.
PromptCCD~\cite{cendra2025promptccd} employs a Gaussian Mixture Model (GMM) as a prompt pool, utilizing a dynamic Gaussian Mixture Prompting (GMP) module to facilitate representation learning and mitigate forgetting.

Unlike traditional classification tasks, NCD does not require immediate labeling of new classes. Instead, \textit{it uses clustering on additional test data for post-hoc evaluation of new class discovery}. This makes current NCD methods \textit{unsuitable} for test-time application, as they rely on further clustering analysis, whereas test-time requires immediate results and the ability to incorporate those results into further training without a separate evaluation phase.

\section{Dataset construction}

We conduct our experiments using three widely recognized benchmark datasets: CIFAR100 (C100)~\cite{cifar100krizhevsky2009learning}, Caltech-UCSD Birds-200-2011 (CUB)~\cite{cubwah2011caltech}, and Tiny ImageNet~\cite{tinyle2015tiny}. Each of these datasets is systematically partitioned into known and unknown classes. The model undergoes training on the known training set and is subsequently evaluated on a mixed set containing both known and unknown classes. Since the primary objective of these datasets is to facilitate new class discovery, we denote the transformed versions as CIFAR100D, CUB-200D, and Tiny-ImageNetD to reflect this adaptation.

The dataset partitioning follows the scheme outlined in Table \ref{tab:dataset}. Specifically, during the training phase, we divide the training set into known and unknown classes based on their class index order. For instance, in CIFAR100, the first 70 classes are designated as known, while the remaining 30 classes are treated as unknown. The supervised training process is then conducted using only the known classes within the labeled training set. More precisely, CIFAR100D consists of classes 0–69 (70 known classes in total), CUB-200D includes classes 0–139 (140 known classes in total), and Tiny-ImageNetD comprises classes 0–139 (140 known classes in total), all of which are utilized for training.

During the test phase, the model is evaluated on the entire unlabeled test set, which includes samples from all categories, enabling new class discovery and classification. While the category labels remain structured according to the original known-unknown splits (e.g., 70+30 for CIFAR100D and 140+60 for CUB-200D and Tiny-ImageNetD), these labels are only used for metric evaluation and are not provided to the model during inference. This setup ensures a realistic scenario for open-world learning, where the model must autonomously identify and categorize previously unseen classes.

\begin{table*}[t]
\centering
\caption{Statistic of the used datasets.}
\resizebox{\linewidth}{!}{
\begin{tabular}{cc|ccc|ccc|ccc}
\toprule
 && \multicolumn{3}{c|}{\textbf{CIFAR100D}} & \multicolumn{3}{c|}{\textbf{CUB-200D}} & \multicolumn{3}{c}{\textbf{Tiny-ImagenetD}} \\ \cmidrule{3-11}
{\textbf{Dataset}} & Labeled   & Known & Unknown & No. of samples     & Known & Unknown  & No. of samples  & Known & Unknown  & No. of samples    \\
\midrule
TrainSet   & $\checkmark$ & 70    & 0 & 42000 & 140    & 0 & 4195 & 140    & 0 & 70000\\
TestSet   & & 70 & 30 & 10000 & 140 & 60 & 5794 & 140 & 60 & 10000 \\
\bottomrule
\end{tabular}
}
\label{tab:dataset}
\end{table*}

\section{Metric Definition}

The evaluation process is structured into two distinct parts: one focusing on known classes and the other on unknown classes. To ensure a comprehensive assessment, we employ both real-time evaluation and post-evaluation strategies.
For test-time evaluation, real-time performance is a critical factor. Thus, we compute and report real-time scores for all evaluation metrics as the model processes each test sample. This approach provides immediate insights into the model’s performance and enables dynamic tracking of classification accuracy and discovery efficiency. 
Alongside these real-time scores, we present the final accumulated values, which represent the overall average performance across the entire test set.
In addition, recognizing that traditional novel class discovery (NCD) methods typically rely on post-evaluation, we also incorporate this approach for comparative analysis. In post-evaluation, all test samples are revalidated collectively after the entire test phase is complete. 
This post-hoc evaluation allows for a more refined assessment by leveraging the full distribution of test samples, potentially improving class assignment and clustering accuracy. By providing both real-time and post-evaluation scores, we ensure a thorough and balanced evaluation of the model’s effectiveness in handling both known and unknown classes.

\subsection{Metrics for known classes}

For the evaluation of known classes, we employ two key metrics to comprehensively assess the model’s performance: Known Accuracy (KA) and Known Forgetting (KF).

\noindent
(1) \textit{Known Accuracy} (\textbf{KA}). 
KA measures the traditional classification accuracy of the model on known classes, reflecting its ability to correctly recognize and classify samples that were part of the training set. This metric serves as a standard benchmark for evaluating the retention of previously learned knowledge:
\begin{equation}
    \text{KA} = \mbb{E}_{c \in \mc{Y}_\text{known}} \frac{1}{|\mc{D}^\text{test}_c|}\sum_{x\in\mc{D}^\text{test}_c}\mb{1}(p(x)=c),
\end{equation}
where $\mc{Y}_\text{known}$ is set of predefined known classes, $\mc{D}^\text{test}_c$ is test samples with ground-truth class $c$, $p(x)$ is the predicted label for sample $x$, $\mb{1}(\cdot)$ is the indicator function (1 if prediction matches true class $c$, 0 otherwise)

\noindent
(2) \textit{Known Forgetting} (\textbf{KF}). KF, on the other hand, quantifies the degree of performance degradation on known classes over time. It captures the extent to which the model forgets previously learned information as it encounters new data, particularly when adapting to novel classes. A lower KF score indicates better knowledge retention, while a higher score suggests significant forgetting.
\begin{equation}
    \text{KF} = \text{KA}_\text{post} - \text{KA}_\text{pre},
\end{equation}
where $\text{KA}_\text{post}$ and $\text{KA}_\text{pre}$ are the KA computed on all test data with known classes, before and after TTD. 

\subsection{Metrics for unknown classes}

For unknown classes, since the predicted label space $\mc{Y}^\text{GT}_\text{seen}$ does not match the cluster label space $\mc{Y}_\text{seen}$, we propose agreement metrics to assess effectiveness. In the test set $\mc{D}^\text{test}$, a sample $x$ has a true label $y\in\mc{Y}^\text{GT}_\text{seen}$ and a predicted cluster label $p(x)\in\mc{Y}_\text{seen}$. We define the subset of $\mc{D}^\text{test}$ with true label $c$ as $\mc{D}^\text{test}_c$, and the cluster with predicted label $p$ as $\mc{C}^\text{test}_p$.

\noindent
(1) \textit{True-label Agreement ratio} (\textbf{TA}). This metric measures the maximum proportion of samples from a given true class that are predicted as the same class:
\begin{equation}
    \text{TA} = \mbb{E}_{c \in \mc{Y}^\text{GT}_\text{seen}} \frac{1}{|\mc{D}^\text{test}_c|}\max_{p\in\mc{Y}_\text{seen}}\left(\sum\nolimits_{x\in\mc{D}^\text{test}_c}\mb{1}(p(x)=p)\right),
\end{equation}
where $\mb{1}(\cdot)$ is the indicator function (1 if true, 0 otherwise).

\noindent
(2) \textit{True-label Entropy} (\textbf{TE}). This metric measures the average entropy $H(\cdot)$ of the predicted labels for samples with that true class:
\begin{equation}
    \text{TE} =\mbb{E}_{c \in \mc{Y}^\text{GT}_\text{seen}}H(\{p(x)|x\in\mc{D}^\text{test}_c\})
\end{equation}
where $H(\cdot)$ is Shannon entropy of predicted label distribution which is used to quantifies uncertainty or diversity in a distribution, $H({z_i}) = -\sum_{z \in \mathcal{Z}} q(z) \log_2 q(z)$.

\noindent
(3) \textit{Cluster Agreement ratio} (\textbf{CA}). This metric measures the maximum proportion of samples from a given predicted cluster that are with the same true label:
\begin{equation}
    \text{CA} = \mbb{E}_{p \in \mc{Y}_\text{seen}} \frac{1}{|\mc{C}^\text{test}_p|}\max_{c\in\mc{Y}^\text{GT}_\text{seen}}\left(\sum\nolimits_{(x,y)\in\mc{C}^\text{test}_p}\mb{1}(y=c)\right).
\end{equation}

\noindent
(4) \textit{Cluster Entropy} (\textbf{CE}). This metric measures the average entropy of the samples that predicted the true class contained in clusters:
\begin{equation}
    \text{CE} =\mbb{E}_{p \in \mc{Y}_\text{seen}}H(\{y|(x,y)\in\mc{C}^\text{test}_p\}).
\end{equation}

\subsection{NCD metrics}

Traditional novel class discovery (NCD) methods typically rely on post-cluster evaluation, where the quality of the discovered clusters is assessed after the entire test set has been processed. To ensure a comprehensive comparison with existing approaches, we also report several widely used clustering evaluation metrics, including Hungarian Cluster Accuracy (HCA)~\cite{meilua2003comparing}, Adjusted Rand Index (ARI)~\cite{rand1971objective}, Normalized Mutual Information (NMI)~\cite{mcdaid2011normalized}, and V-Measure~\cite{rosenberg2007v}.
Note that these metrics are only evaluated after TTD, say post evaluation.

\noindent
(1) \textit{Hungarian Cluster Accuracy} (\textbf{HCA}).
This metric measures the clustering accuracy by computing an optimal one-to-one mapping between predicted clusters and ground-truth labels using the Hungarian algorithm. It provides an intuitive evaluation of how well the discovered clusters align with the actual class distributions. HCA can be computed as
\begin{equation}
    \text{HCA} = \mbb{E}_{(x,y)\in\mc{D}^\text{test}} (y = \text{map}(p(x))),
\end{equation}
where $\text{map}(\cdot)$is the optimal mapping from clustering to true labels obtained based on the Hungarian algorithm

\noindent
(2) \textit{Adjusted Rand Index} (\textbf{ARI}).
ARI quantifies the similarity between the predicted clustering assignments and the ground-truth labels while adjusting for chance. It accounts for both correct pairwise clustering and misclustered pairs, offering a robust measure of clustering consistency.
\begin{equation}
    \text{ARI} = \frac{\text{RI} - \mbb{E}[\text{RI}]}{\max(\text{RI}) - \mbb{E}[\text{RI}]},
\end{equation}
where Rand Index $(\text{RI}) = \frac{a + b}{C^2_n}$, $a$ is the logarithm of samples of the same class assigned to the same cluster, and $b$ is the logarithm of samples of different classes assigned to different clusters. $n$ is the total number of samples, combination $C^2_n = \frac{n(n-1)}{2}$ and $\mbb{E}[\text{RI}]$ is the expected value of RI.

\noindent
(3) \textit{Normalized Mutual Information} (\textbf{NMI}). NMI assesses the mutual dependence between predicted and true labels by measuring the shared information between the two distributions. A higher NMI value indicates better alignment between the discovered clusters and the actual categories. The value interval of NMI is [0,1], and a larger value indicates a higher degree of information sharing between the clustering results and the real labels.
\begin{equation}
\text{NMI}(\mc{U},\mc{V}) = \frac{2 \cdot I(\mc{U};\mc{V})}{H(\mc{U}) + H(\mc{V})},
\end{equation}
where $\mc{U}$ is collection of true labels and $\mc{V}$ is collection of predictions.
$I(\mc{U};\mc{V})$ is Mutual Information where $I(\mc{U};\mc{V})=H(\mc{U})-H(\mc{U}|\mc{V})$. 
$H(\mc{U})$ is the entropy of true label, ${H(\mc{U})} = -\sum_{c = 1}^C {p(c)\log p(c)}$, and $H(\mc{V})$ is the entropy of prediction, ${H(\mc{V})} = -\sum_{k = 1}^K {p(k)\log p(k)}$.

\noindent
(4) \textit{V-Measure} (\textbf{VM}). The VM is taken in the interval [0,1], which simultaneously constrains the purity and coverage of the clusters through the harmonic mean. Both VM and NMI are symmetric metrics that support the comparison of clusters and categories at different scales.
\begin{equation}
\text{V-Measure} = \frac{2 \cdot h \cdot c}{h + c},
\end{equation}
where homogeneity $h=1-\frac{H(\mc{U}|\mc{V})}{H(\mc{U})}$, and completeness $c=1-\frac{H(\mc{V}|\mc{U})}{H(\mc{V})}$. $H(\mc{U}|\mc{V}) = - \sum_{k = 1}^K {\sum_{t = 1}^T {p(k,t)\log \frac{{p(k,t)}}{{p(k)}}} }$ and $H(\mc{V}|\mc{U}) = -\sum_{t = 1}^T {\sum_{k = 1}^K {p(t,k)\log \frac{{p(t,k)}}{{p(t)}}}}$, where $p(t) = \frac{N_t}N$ is the sample proportion of class $t$, $p(k) = \frac{N_k}N$ is the sample proportion of cluster $k$, and $p(t,k)=\frac{count_k(t)}N$ is the joint distribution probability.

\section{Hyper-parameter analysis}

\subsection{Number of selected neighboring buckets in graph-based neighbor searching}

\begin{figure}[h]
\centering
\includegraphics[width=\linewidth]{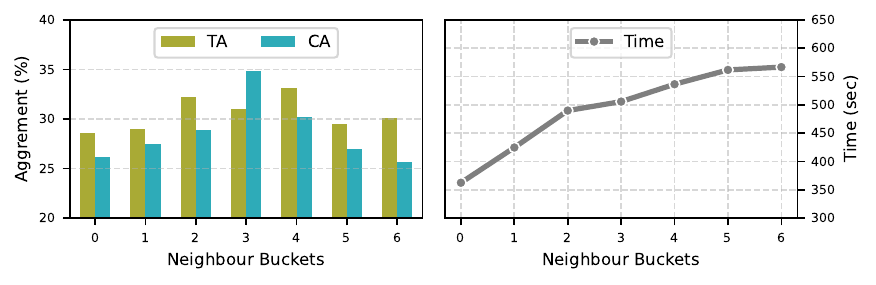}
\vspace{-10px}
\caption{Comparisons of different numbers of compared samples in buckets}
\label{fig:neighbour}
\end{figure}

In Fig.~\ref{fig:neighbour}, we present a comparative analysis of the impact of varying the number of neighbor buckets in our method during TTD. The choice of bucket size directly influences both the effectiveness of new class discovery and computational efficiency.
When the number of buckets is small, the model has access to a limited number of nearest neighbors. This scarcity of reference samples hinders the model’s ability to effectively discover and categorize novel classes, as the available nearest-neighbor information may be insufficient to form meaningful clusters. However, a smaller bucket size also results in lower computational overhead, as fewer comparisons are required during the clustering process.
Conversely, when the number of buckets is large, the model can consider a greater number of nearest neighbors. While this increases the availability of reference samples for new class discovery, an excessive number of neighbors can introduce noisy or misleading information, leading to incorrect clustering assignments. Additionally, the increased volume of sample comparisons significantly raises computational costs, prolonging processing time.
Thus, selecting an optimal number of neighbor buckets is a trade-off between discovery accuracy and computational efficiency. A balanced choice ensures that the model captures sufficient nearest-neighbor information while minimizing erroneous interferences and excessive time consumption.

\subsection{Number of compared samples in buckets}

After retrieving the associated bucket using the hash value, we calculate the distances between the samples within the bucket and the target test sample. The top-$k$ nearest samples are then identified, and their labels are used for voting to determine the final classification of the target test sample. Thus, selecting an appropriate number of voting samples, i.e., determining the optimal value of $k$, is a critical factor in achieving accurate classification.
To investigate the impact of different $k$ values, we conduct comparative experiments with varying settings, as presented in Table \ref{tab:top-k}. The results reveal that when $k$ is relatively small, TA decreases, whereas CA improves. This suggests that samples from the same class are more likely to be concentrated within known classes, while the number of correctly captured samples in new class clusters remains limited.
On the other hand, as $k$ increases, both CA and TA decline, indicating that the internal structure of new class clusters becomes more confused. This implies that an excessively large $k$ introduces more noise, making it harder for the model to differentiate between novel categories.
These findings underscore the importance of carefully selecting an appropriate $k$ value to maintain effective discovery.

\begin{table}[h]
\centering
\caption{Different number of compared samples in buckets on CIFAR100D.}
\resizebox{\linewidth}{!}{
\begin{tabular}{c|cccc|ccccc}
\toprule
\multirow{2}{*}{} & \multicolumn{4}{c|}{\textbf{Real-time Eval}} & \multicolumn{5}{c}{\textbf{Post Eval}}       \\ \cmidrule{2-10} 
                  & TA       & TE     & CA      & CE     & TA    & TE   & CA    & CE   & KF    \\
\midrule
0                   & 19.82                                            & 1.01    & 39.27    & 2.92   & 22.76 & 1.22 & 25.80 & 1.65 & -6.22 \\
5                   & 18.05                                            & 0.72    & 56.06    & 1.40   & 35.26 & 1.34 & 27.83 & 1.70 & -7.10 \\
10                  & 21.11                                            & 0.66    & 56.87    & 1.27   & 31.03 & 1.07 & 34.81 & 1.50 & -3.47 \\
15                  & 16.73                                            & 0.75    & 52.15    & 1.34   & 30.40 & 1.50 & 28.67 & 1.54 & -4.43 \\
20                  & 13.79                                            & 0.80    & 54.24    & 1.37   & 29.23 & 1.52 & 26.15 & 1.60 & -6.32  \\
\bottomrule
\end{tabular}
}
\label{tab:top-k}
\end{table}

\subsection{Number of random directions when hashing features}

\begin{figure}[h]
\centering
\includegraphics[width=\linewidth]{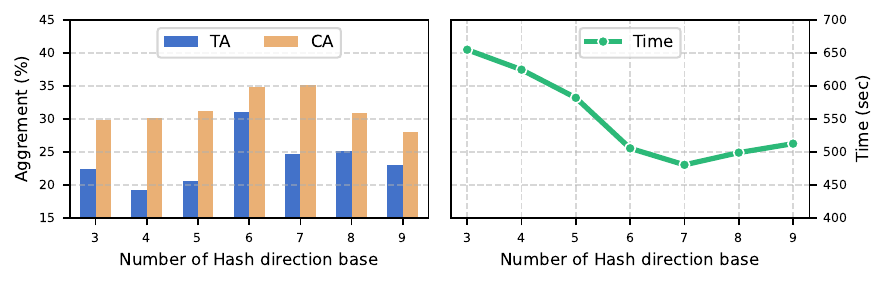}
\caption{Metrics of different Hash angles}
\label{fig:directionnumer}
\end{figure}

When constructing the hash memory, we utilize multiple random directional bases to define feature orientations and partition the angular space. The number of these bases plays a crucial role in performance. To investigate its impact, we conduct comparative experiments with different base quantities, as shown in Fig.~\ref{fig:directionnumer}.
The results indicate that when the number of bases is too low, the angular space is insufficiently separated, leading to decreased TA and CA. Additionally, the reduced ability to distinguish between different feature directions increases comparison time, further affecting efficiency.
On the other hand, when the number of bases is too high, we observe a similar decline in performance. This suggests that an excessive number of buckets does not improve sample-level discrimination. Instead, it increases the complexity of searching within nearest-neighbor buckets, making comparisons less effective and computationally more demanding.
These findings emphasize the importance of choosing an optimal number of random directional bases to achieve a balanced trade-off between discovery effectiveness, and computational efficiency.

\subsection{Factor for exponential moving average (EMA)}

Although our method does not update the model itself, it continuously refines the prototype representations of each class. To assess the impact of prototype updates on both known and seen classes, we conduct experiments, with the results presented in Tables~\ref{tab:novelema} and \ref{tab:oldema}.
The findings indicate that an overly aggressive update of newly discovered class prototypes amplifies errors introduced during the discovery process, leading to a decline in TA.
For known classes, the most effective approach is to maintain static prototypes without updates. Updating these prototypes not only exacerbates catastrophic forgetting but also increases confusion between known and unknown classes, further deteriorating overall performance.

\begin{table}[h]
\centering
\caption{Comparisons of different EMA factors for unknown classes.}
\resizebox{\linewidth}{!}{
\begin{tabular}{c|cccc|ccccc}
\toprule
 & \multicolumn{4}{c|}{\textbf{Real-time Eval}} & \multicolumn{5}{c}{\textbf{Post Eval}}       \\ \cmidrule{2-10} 
$\alpha$       & TA       & TE     & CA      & CE     & TA    & TE   & CA    & CE   & KF    \\
\midrule
1.0 & 14.55 & 0.63 & 65.83 & 1.02 & 5.17  & 0.33 & 44.31 & 0.95 & -0.71 \\
0.99       & 14.57 & 0.59 & 66.91 & 1.03 & 22.80 & 0.83 & 36.92 & 1.16 & -2.59 \\
0.9 & 21.11 & 0.66 & 56.87 & 1.27 & 31.03 & 1.07 & 34.81 & 1.50 & -3.47 \\
0.8 & 21.07 & 0.69 & 55.75 & 1.25 & 25.30 & 1.12 & 31.57 & 1.60 & -3.26 \\
0.7 & 19.23 & 0.71 & 57.19 & 1.26 & 16.27 & 1.09 & 29.18 & 1.75 & -2.43 \\
0.6 & 19.74 & 0.75 & 56.43 & 1.23 & 16.97 & 1.16 & 27.42 & 1.76 & -2.74 \\
0.5 & 19.68 & 0.70 & 58.83 & 1.18 & 13.27 & 0.96 & 32.16 & 1.71 & -2.00  \\
\bottomrule
\end{tabular}
}
\label{tab:novelema}
\end{table}

\begin{table}[h]
\centering
\caption{Comparisons of different EMA factors for known classes.}
\resizebox{\linewidth}{!}{
\begin{tabular}{c|cccc|ccccc}
\toprule
 & \multicolumn{4}{c|}{\textbf{Real-time Eval}} & \multicolumn{5}{c}{\textbf{Post Eval}}       \\ \cmidrule{2-10} 
$\alpha$       & TA       & TE     & CA      & CE     & TA    & TE   & CA    & CE   & KF    \\
\midrule
1.0 & 21.11 & 0.66 & 56.87 & 1.27 & 31.03 & 1.07 & 34.81 & 1.50 & -3.47  \\
0.99 & 16.55 & 0.65 & 60.05 & 1.17 & 19.13 & 1.13 & 30.60 & 1.56 & -3.47  \\
0.9             & 14.54 & 0.59 & 59.65 & 1.23 & 10.53 & 0.65 & 32.06 & 1.40 & -2.76  \\
0.8             & 16.19 & 0.60 & 55.21 & 1.37 & 14.77 & 0.77 & 32.50 & 1.43 & -4.71  \\
0.7             & 19.51 & 0.66 & 47.82 & 1.58 & 17.07 & 0.90 & 29.33 & 1.45 & -16.49 \\
0.6             & 17.93 & 0.64 & 47.34 & 1.58 & 19.83 & 0.96 & 25.69 & 1.44 & -22.33 \\
0.5             & 19.04 & 0.66 & 48.74 & 1.61 & 19.90 & 1.15 & 20.99 & 1.39 & -25.69  \\
\bottomrule
\end{tabular}
}
\label{tab:oldema}
\end{table}

\section{Different thresholds for thresholding methods}

\begin{figure}[h]
\centering
\includegraphics[width=\linewidth]{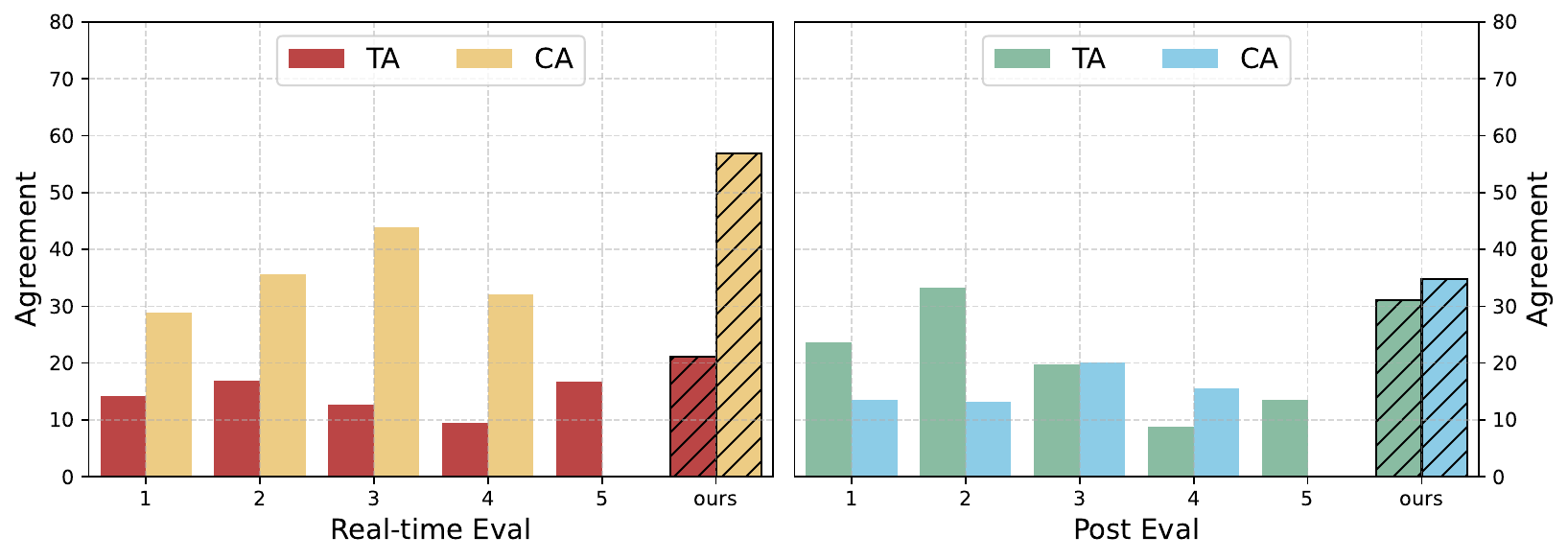}
\caption{Euclidean threshold}
\label{fig:Euclidean}
\end{figure}

\begin{figure}[h]
\centering
\includegraphics[width=\linewidth]{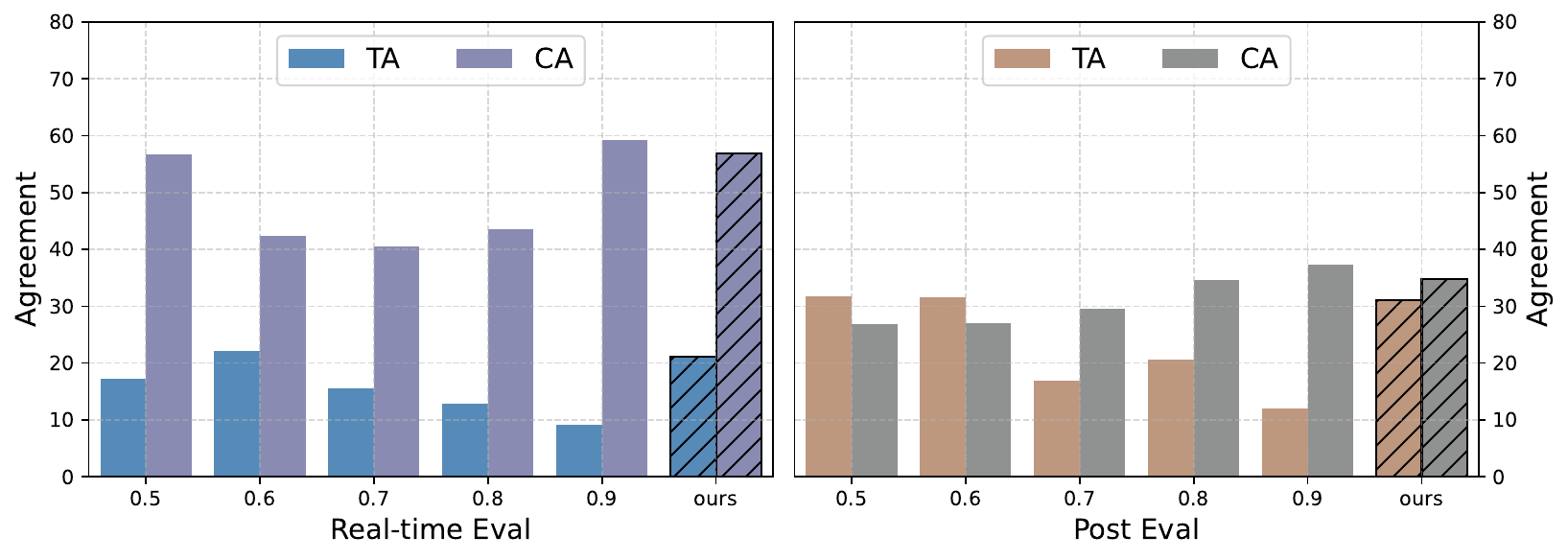}
\caption{Cosine threshold}
\label{fig:Cosine}
\end{figure}

\begin{figure}[h]
\centering
\includegraphics[width=\linewidth]{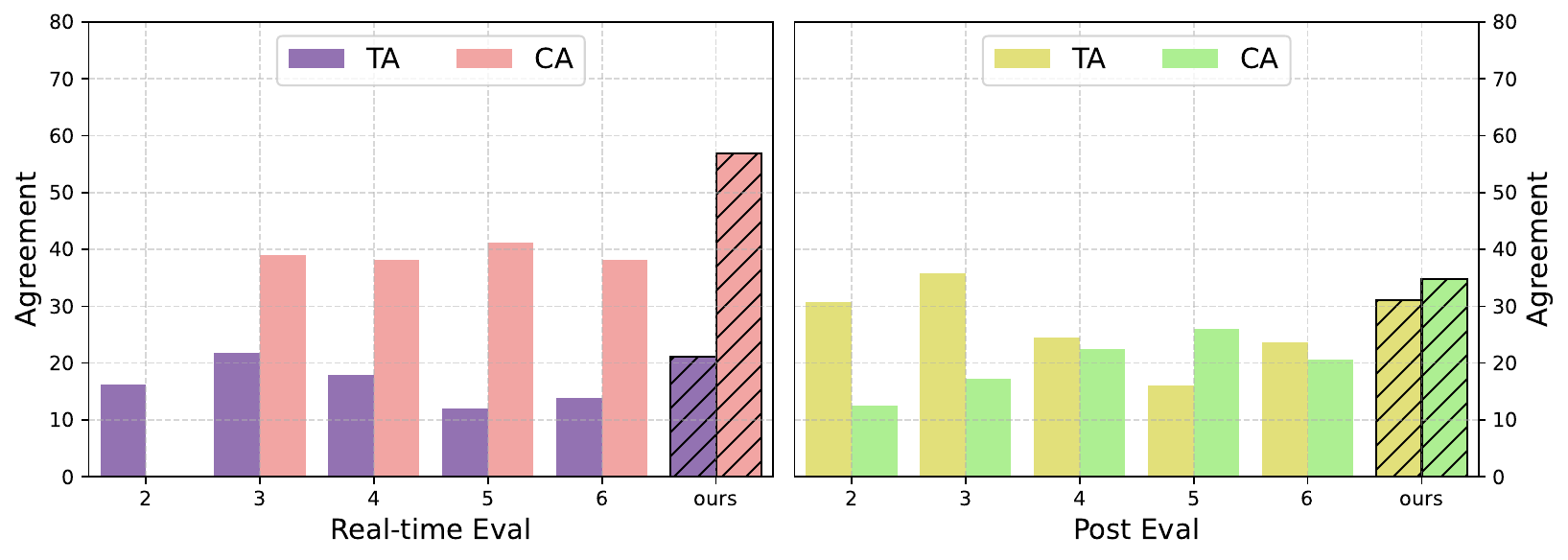}
\caption{Magnitute threshold}
\label{fig:Mag}
\end{figure}

\begin{figure}[h]
\centering
\includegraphics[width=\linewidth]{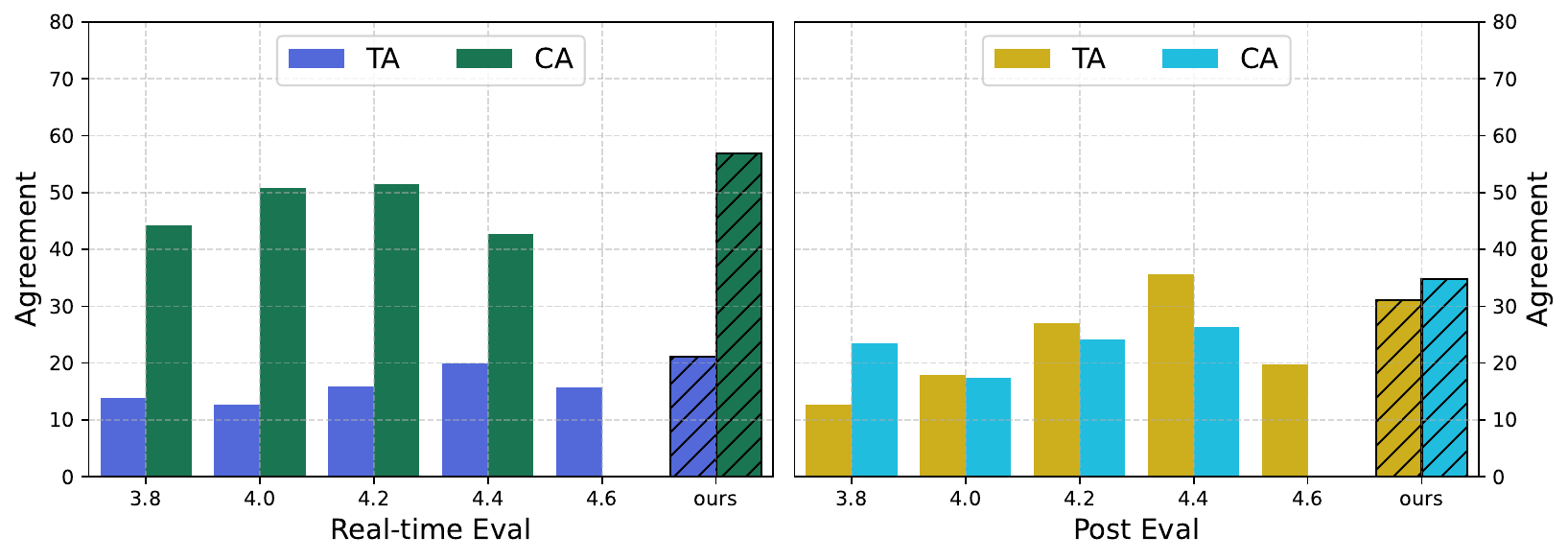}
\caption{Entropy threshold}
\label{fig:Entropy}
\end{figure}

In our main text, we compare several methods that utilize threshold-based approaches, for which we provide an optimized threshold selection.
In Figs.~\ref{fig:Euclidean}, \ref{fig:Cosine}, \ref{fig:Mag} and \ref{fig:Entropy}, we present a comparison of threshold selection across different methods alongside our results. The findings indicate that competing methods are highly sensitive to threshold choices, making it difficult to achieve both high TA and high CA simultaneously.
In contrast, our method maintains a better balance between TA and CA, demonstrating greater stability and robustness across different threshold settings.

\section{Running Time}

\begin{figure}[h]
\centering
\includegraphics[width=.5\linewidth]{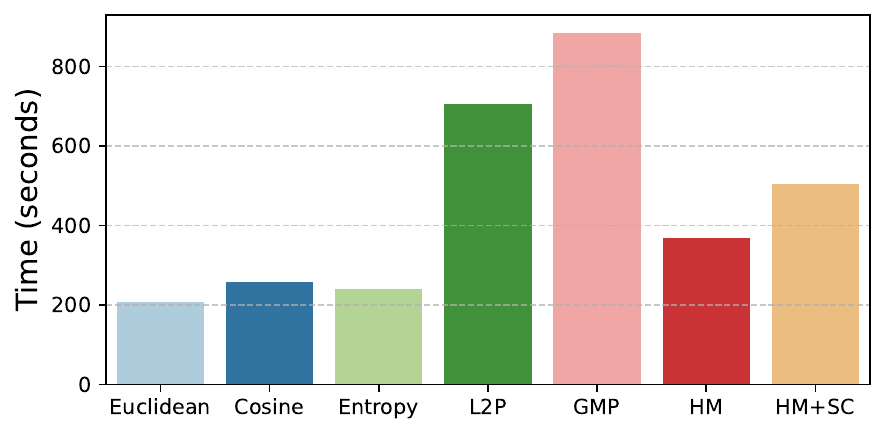}
\caption{Comparisons of running time.}
\label{fig:running_time}
\end{figure}

We evaluate the running time of the compared methods.
The results are shown in Fig.~\ref{fig:running_time}, which show that training-required methods require significantly more time, while training-free approaches are generally faster. Our method takes longer than threshold-based methods but remains faster than training-based approaches. Additionally, incorporating the SC strategy further increases running time.

\end{document}